\documentclass[10pt,twocolumn,letterpaper]{article}

\usepackage{iccv}              
\usepackage[linesnumbered,ruled,lined]{algorithm2e}
\usepackage{multirow}
\usepackage{colortbl}
\usepackage{color}
\usepackage{booktabs}
\usepackage{graphicx}
\usepackage{pifont}
\usepackage{bbding}
\usepackage{colortbl}
\usepackage{color}

\newcommand{\Ra}[1]{\textcolor{LimeGreen}{\textbf{#1}}} 
\newcommand{\Rb}[1]{\textcolor{VioletRed}{\textbf{#1}}} 
\newcommand{\Rc}[1]{\textcolor{SkyBlue}{\textbf{#1}}} 
 
\def\YES{\textcolor{SeaGreen}{\ding{51}}}
\def\NO{\textcolor{Red}{\ding{55}}}

\definecolor{iccvblue}{rgb}{0.21,0.49,0.74}
\usepackage[pagebackref,breaklinks,colorlinks,allcolors=iccvblue]{hyperref}

\def\paperID{7048} 
\def\confName{ICCV}
\def\confYear{2025}

\title{Learning to Generalize without Bias for Open-Vocabulary Action Recognition}

\author{
Yating Yu$^{1}$\thanks{Equal Contribution} \quad
Congqi Cao$^{1\ast}$\thanks{Corresponding Author} \quad
Yifan Zhang$^{2}$ \quad
Yanning Zhang$^1$ \\
$^1$Northwestern Polytechnical University \quad
$^2$Institute of Automation, Chinese Academy of Sciences \\
{\tt \small yatingyu@mail.nwpu.edu.cn, congqi.cao@nwpu.edu.cn,} \\
{\tt \small yfzhang@nlpr.ia.ac.cn, yanningzhang@nwpu.edu.cn}
}

\begin{document}
\maketitle
\begin{abstract}
Leveraging the effective visual-text alignment and static generalizability from CLIP, recent video learners adopt CLIP initialization with further regularization or recombination for generalization in open-vocabulary action recognition in-context.
%
However, due to the static bias of CLIP, such video learners tend to overfit on shortcut static features, thereby compromising their generalizability, especially to novel out-of-context actions.
%
To address this issue, we introduce \textbf{Open-MeDe}, a novel \underline{Me}ta-optimization framework with static \underline{De}biasing for \underline{Open}-vocabulary action recognition.
From a fresh perspective of generalization, Open-MeDe adopts a meta-learning approach to improve ``\textbf{known-to-open generalizing}'' and ``\textbf{image-to-video debiasing}'' in a cost-effective manner.
Specifically, Open-MeDe introduces a cross-batch meta-optimization scheme that explicitly encourages video learners to quickly generalize to arbitrary subsequent data via virtual evaluation, steering a smoother optimization landscape.
In effect, the free of CLIP regularization during optimization implicitly mitigates the inherent static bias of the video meta-learner.
We further apply self-ensemble over the optimization trajectory to obtain generic optimal parameters that can achieve robust generalization to both in-context and out-of-context novel data.
Extensive evaluations show that Open-MeDe not only surpasses state-of-the-art regularization methods tailored for in-context open-vocabulary action recognition but also substantially excels in out-of-context scenarios.
Code is released at \href{https://github.com/Mia-YatingYu/Open-MeDe}{https://github.com/Mia-YatingYu/Open-MeDe}.
\end{abstract}

\section{Introduction}
\label{sec:intro}
\begin{figure}[t]
    \centering
    \includegraphics[width=\columnwidth]{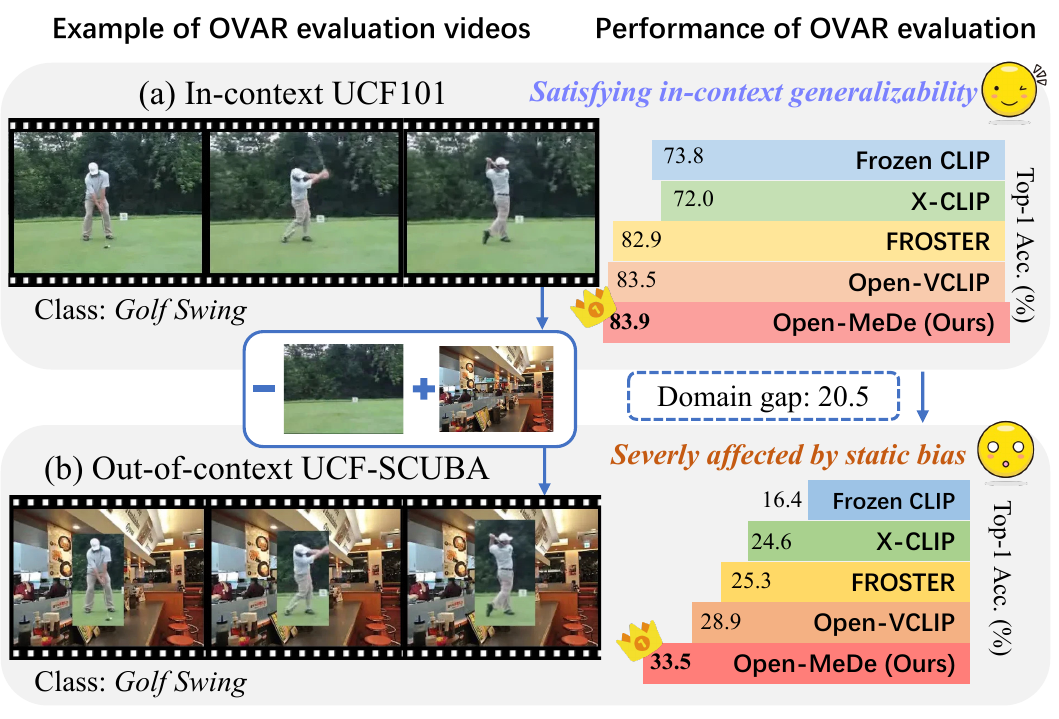}
    \caption{Performance comparison (Top-1 Acc (\%)) under various open-vocabulary evaluation settings where the video learners except for CLIP are tuned on Kinetics-400~\cite{k400} with frozen text encoders. The satisfying in-context generalizability on UCF101~\cite{UCF101} (a) can be severely affected by static bias when evaluating on out-of-context UCF-SCUBA~\cite{li2023mitigating} (b) by replacing the video background with other images.}
    \label{fig:teaser}
\end{figure}
Open-vocabulary action recognition (OVAR) aims to identify test videos whose classes are not previously encountered during the training phase, which challenges the generalization and zero-shot capabilities of the video learners~\cite{zhu2023orthogonal,brattoli2020rethinking,wu2023revisiting,wang2023visual}. 
Recently, the emergence of image-based visual-language (I-VL) pre-training, such as CLIP~\cite{radford2021learning} and ALIGN~\cite{jia2021scaling}, has shown promising zero-shot inference in image-based tasks. 
Inspired by this success, recent attempts~\cite{brattoli2020rethinking,wasim2023vita,chen2024ost,ni2022expanding,pan2022st,cao2024scene} have been made to adapt CLIP for general action recognition via additional temporal modeling following the ``\textit{pre-train, prompt and fine-tune}'' paradigm~\cite{wang2021actionclip}. 
Broadly, these video learners optimize the learnable parameters from the start point of CLIP, pursuing decent performance on the training videos, known as standard fine-tuning objectives.
However, adapting CLIP to the video domain, especially for OVAR, is extremely challenging, as the video learners with standard fine-tuning objectives often lead to overfitting, which achieves improved specialization at the cost of generalization degradation.

To build an improved zero-shot video learner, Open-VCLIP~\cite{wu2024building} and FROSTER~\cite{huang2024froster} propose to regularize the fine-tuning process curbing deviation from CLIP's generalization from the perspective of model patching~\cite{ilharco2022patching} and knowledge distillation~\cite{castro2022fitclip,dai2022enabling,pei2023clipping}, respectively. 
In~\cref{fig:teaser}, these methods have achieved satisfying performance compared to frozen CLIP and X-CLIP~\cite{ni2022expanding} on UCF101~\cite{UCF101} dataset under in-context open-vocabulary evaluation, where the action categories have strong correlations with the context in videos.
However, when it comes to the out-of-context evaluation in SCUBA~\cite{li2023mitigating}, where the video background is replaced by other images, the performance degrades severely.
As these video learners are intimately tied to the learning of shortcut static features, which manifest as static bias, they interfere with the learning of motion cues, resulting in poor out-of-context generalization~\cite{duan2022mitigating}.
Based on these observations, we argue that the static generalization of CLIP can (1) effectively adapt to in-context scenarios for OVAR by regularizing video learners; yet (2) it undesirably hinders the sensitivity of such video learners to motion cues, exerting a notable detrimental impact on generalization under out-of-context, open-vocabulary setting.

\textit{How can we encourage the emergence of such robust open-vocabulary generalization for both in-context and out-of-context scenarios?}
We explore an explicit approach to this problem: as the video learner is trained with a sampled batch of videos at each gradient step, our objective is to optimize the learner from a meta-learning standpoint so that it can quickly adapt to arbitrary subsequent data, thereby minimizing inherent biases toward known data and static cues.

Based on this insight, we propose \textbf{Open-MeDe}, the first \underline{Me}ta-learning based framework with static \underline{De}biasing for in-context and out-of-context \underline{Open}-vocabulary action recognition.
Meta-learning, also known as ``\textit{learning to learn}'', incorporates virtual evaluation during the training process for better generalization~\cite{finn2017model,nichol2018first,antoniou2018train}.
In our meta-learning scheme, the ``\textit{learning to generalize}'' process is enhanced by naturally treating sequences of adjacent batches sampled from the training set as a distribution of tasks. 
More concretely, our procedure optimizes the video learner to obtain fast weights by gradient descent updates on the current batch (\ie, \textit{meta training}), while evaluating the subsequent batch (\ie, \textit{meta testing}) based on fast weights of the learner, which mimics a known-to-open task.
Based on the evaluation performance in \textit{meta testing}, our procedure can further optimize the learner to obtain more generalizable video-specific knowledge against inherent known and static biases.
In effect, this cross-batch meta-optimization formulates a meta-learner free of CLIP regularization, thereby facilitating smoother optimization and robust video representation learning for fast known-to-open generalizing, thus enhancing image-to-video debiasing.
Tailored to the optimization trajectory of the video learner, we further employ self-ensemble stabilization, \ie, Gaussian Weight Average (GWA), to derive generic optima for robust generalization at open-vocabulary test time.
Overall, while integrating the same video learner, our model-agnostic Open-MeDe outperforms existing regularization-based methods, which strikes a promising balance on in-context and out-of-context generalization settings (\cref{fig:teaser}).

The contribution of our work can be summarized as:
\begin{itemize}
    \item We introduce a novel meta-learning based framework, Open-MeDe, which provides new insights for more generalized open-vocabulary action recognition.

    \item We propose cross-batch meta-optimization and self-ensemble stabilization, which effectively power known-to-open generalizing and image-to-video debiasing of the video learner for robust generalizability.

    \item We conduct extensive evaluations on various scenarios including base-to-novel, cross-dataset, and out-of-context open-vocabulary action recognition. Experimental results show that Open-MeDe consistently improves performance across all the benchmarks.

\end{itemize}
\begin{figure*}[t]
    \centering
    \includegraphics[width=\textwidth]{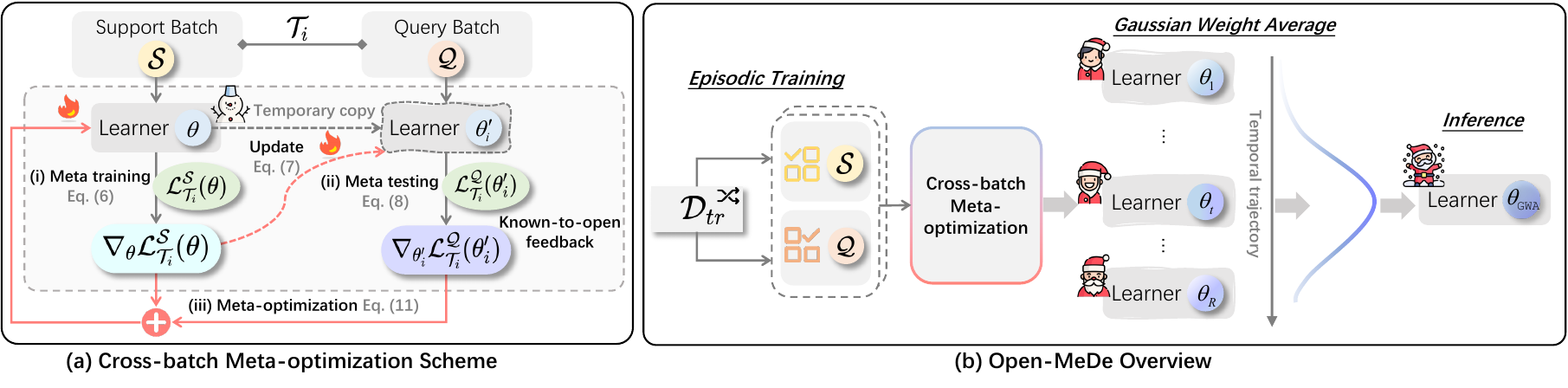}
    \caption{Illustration of our framework. (a) The cross-batch meta-optimization scheme aims to mimic the known-to-open generalization task $\mathcal{T}_i$ by performing the gradient descent update (\ie, \textit{meta training}) on the support batch $\mathcal{S}$ and virtual evaluation (\ie, \textit{meta testing}) on the query batch $\mathcal{Q}$. Then, the video learner is optimized by both class-specific losses from $\mathcal{S}$ and task feedback from $\mathcal{Q}$ for more generalizable knowledge against inherent known and static biases. (b) Overview of the Open-MeDe framework with self-ensemble stabilization. During the episodic training process, we exploit the optimization trajectory of the video learner to perform Gaussian Weight Average (GWA) to derive generic optima for robust generalization.}
    \label{fig:overview}
\end{figure*}
\section{Related Work}
\label{sec:related}
\subsection{Adapting CLIP to Action Recognition}
A seminal work of I-VL,  CLIP~\cite{radford2021learning} has demonstrated remarkable static generalization, achieving promising performance in image-based zero-shot inference. 
Despite extensive works~\cite{rasheed2023fine,wang2021actionclip,wu2023revisiting} fully fine-tuning the video learner, a collection of studies focuses on adopting lightweight adapters~\cite{yang2023aim,pan2022st,cao2024task} or incorporating learnable prompts~\cite{wasim2023vita,ju2022prompting}  for easy video adaptation.
However, these video learners adhere to the standard fine-tuning paradigm, which tends to overfit in the closed-set setting, thereby limiting expertise in open-vocabulary settings.
To this end, Open-VCLIP~\cite{weng2023open} regularizes the fine-tuning process of the video learner, preventing deviation from CLIP's generalization, by interpolating frozen CLIP weights with the current learner on the fly.
FROSTER~\cite{huang2024froster} and STDD~\cite{yu2025building} enforce the regularization from the perspective of knowledge distillation~\cite{chen2022improved,romero2014fitnets,castro2022fitclip,deng2021comprehensive}, aligning features of the video learner and frozen CLIP via a tailored residual module.
Despite demonstrating superiority in open-vocabulary evaluations,  the increased computational overhead and excessive reliance on static cues introduced by CLIP regularization hinder efficient adaptation and robust generalization.
In contrast, we approach the problem of adapting CLIP-based video learners to OVAR from a fresh view of “learning to generalize without bias”. During training, the learner is explicitly forced to quickly generalize to forthcoming data by sorely resorting to the knowledge learned by itself rather than by the virtue of CLIP’s static generalization.

\subsection{Meta-learning}
Rather than directly learning from experiences, with the goal of learning to learn, meta-learning can quickly generalize to new tasks by leveraging prior learning abilities~\cite{hospedales2021meta}.
As the representative works in meta-learning, MAML~\cite{finn2017model} boasts simplicity and has actively driven the development of the gradient-based methods in few-shot learning. 
Recently, meta-learning techniques have also been explored in zero-shot learning~\cite{verma2020meta,park2024prompt,liu2021task,huang2019generative} and domain adaptation~\cite{li2018learning}, which typically perform episode-wise training by dividing the training set into support and query sets with different classes distributions.
Targeting long-tailed issues within closed-set video scene generation, MVSGG~\cite{xu2022meta} employs meta-learning across several manually predefined task types, which are partitioned based on specific conditional biases in the training data.
However, these approaches are often prone to meta-overfitting due to insufficient meta tasks and limited application scopes of generalization.
Differently, our work tackles ubiquitous challenges in video understanding beyond closed-set and in-context settings, \ie, mitigating static bias of video learners for open-vocabulary generalization. 
To the best of our knowledge, we are the first to directly integrate the mini-batch training mechanism with meta-learning to naturally mimic diverse known-to-open tasks utilizing cross-batch data without additional computational overhead. 

\section{Method}
\label{sec:method}
\subsection{Preliminaries}
\noindent {\bf Action recognition with CLIP-based video learner.}
%
%
Consider a CLIP-based video learner with a ViT architecture~\cite{dosovitskiy2020image}, that incorporates temporal modeling for video understanding~\cite{wu2023revisiting,wang2021actionclip,weng2023open,zhu2023orthogonal,yang2023aim,wasim2023vita}.
Next, we present the standard vision-only fine-tuning paradigm that applies such a video learner $f_{\theta_v}$ with a frozen text encoder $f_{\theta_t}$ to action recognition. 
Specifically, given a video clip $V_i$, and a candidate action label $T_j\in \mathcal{Z}_{tr}$ described in predefined textual templates (\eg, ``\textit{a video of \{action\}}'') from the training set $\mathcal{D}_{tr}$, the similarity is calculated as:
\begin{equation}
    s_{i,j}=\frac{\left \langle v_i,t_j \right \rangle }{\left \| v_i \right \| \left \| t_j \right \| },v_i=f_{\theta_v}(V_i),t_j=f_{\theta_t}(T_j),
\end{equation}
where the training objective is to maximize it of the matched $V_i$ and $T_j$, or to minimize it otherwise. The loss function is implemented by the cross-entropy loss in~\cite{wu2023revisiting,chen2021empirical,radford2021learning} as:
\begin{equation}
\label{eq:ce}
    \mathcal{L}_{CE} = -\frac{1}{B}\sum_{i}^{B}\sum_{k}^{K}y_{i,k}\log \left ( \frac{\exp(s_{i,k})}{\sum_{j}^{K}\exp (s_{i,j})}  \right ),  
\end{equation}
where $B$ and $K$ denote the minibatch size and the number of all known classes, respectively. If the $i$-th video belongs to the $k$-th class, $y_{i,k}$ equals $1$; otherwise, $y_{i,k}$ equals $0$.
In OVAR, the trained video learner should achieve good generalization on test data with the class label $T_i\in \mathcal{Z}_{te}$, where $\mathcal{Z}_{te} \cap \mathcal{Z}_{tr}=\emptyset $.

\noindent {\bf Model-agnostic meta-learning (MAML).}
MAML~\cite{finn2017model} is a gradient-based meta-optimization framework designed for few-shot learning, which aims to learn good initialization such that a few gradient steps will lead to fast learning on new tasks.
Formally, consider a model $f_\theta$ with parameters $\theta$, MAML learns a set of initial weight values, which will serve as a good starting point for fast adaptation to a new task $\mathcal{T}_i$, sampled from a task distribution $p(\mathcal{T})$.
When adapting to the task $\mathcal{T}_i$, the fast weights $\theta_i'$ are computed \textit{w.r.t.} examples from $\mathcal{T}_i$ though single inner-loop update as:
\begin{equation}
    \theta_i'=\theta-\alpha \nabla_\theta \mathcal{L}_{\mathcal{T}_i}(f_\theta),
\end{equation}
where $\alpha$ denotes the step size for inner loops. Then, the model with fast weights $f_{\theta_i'}$ is evaluated on new samples from the same task $\mathcal{T}_i$, to act as the feedback (\ie, loss gradients) to adapt to current task $\mathcal{T}_i$ to optimize the initialization $\theta$ for generalization as:
\begin{equation}
\label{eq:maml}
    \theta \leftarrow \theta-\beta \nabla_\theta \sum_{\mathcal{T}_i}\mathcal{L}_{\mathcal{T}_i}(f_{\theta_i'}).
\end{equation}
where $\beta$ is the step size for outer loops. 
Computationally, due to the additional backward propagation burden of the gradient by gradient update, MAML presents a first-order approximation, FOMAML, by dropping the backward pass. 

\subsection{Open-MeDe}
As discussed above, the standard fine-tuning paradigm can cause the video learner to overfit to the known classes during training, leading to poor zero-shot capabilities. Also, CLIP regularization-based approaches face challenges in achieving robust generalization due to the excessive reliance on superficial static cues in videos.
To tackle these issues, we draw upon the philosophy and methodology from meta-learning, and propose Open-MeDe framework, which is illustrated in~\cref{fig:overview}, to enhance both know-to-open generalizing and image-to-video debiasing simultaneously.

\subsubsection{Cross-batch meta-optimization}
Our Open-MeDe framework primarily adopts a cross-batch meta-optimization scheme (in~\cref{fig:overview}(a)) to enhance the video learner via \textit{meta training and testing}, enabling it to acquire generalizable, video-specific knowledge instead of overly exploiting static biases. 
Note that we neither sample from a distribution of $N$-way $K$-shot tasks as done in few-shot MAML nor deliberately split the training set into support and query sets as Meta-ZSL~\cite{liu2021task,verma2020meta} suggested. Instead, our support and query examples are constructed effortlessly and arbitrarily by the default training data sampler.
In effect, we consider this arbitrariness a blessing for building the natural ``\textit{known-to-open generalization task}'', since the known biases in \textit{meta training} data do not hold in \textit{meta testing} data due to different inherent label distributions across batches.
A known-to-open task can be created by extending the original gradient step into two consecutive mini-batches in one pass, with the current batch acting as support data and the subsequent batch as query data.
Specifically, in line with the episode-wise training akin to MAML, we first train the learner within an inner loop (\ie, \textit{meta training}), where the fast weights are obtained through a single gradient step for each support batch.
Following this adaptation, in the outer loop, query videos are sampled to evaluate the generalization performance of the adapted learner with fast weights (\ie, \textit{meta testing}).
In this work, our framework further updates the fast weights of the learner based on the evaluation performance during \textit{meta testing}, which then provides feedback for the task to derive more generalizable optimization for the learner.

\noindent {\bf Meta training.}
At each training iteration, we first utilize each support batch $\mathcal{S}=\{V_i,T_i\}^B$ from the task $\mathcal{T}_i$ to train the video learner $f_{\theta}$ (with parameters $\theta$), via one standard gradient step. The inner loop update is governed by the loss on the support batch as:
\begin{equation}
\label{eq:spt}
     \mathcal{L}_{\mathcal{T}_i}^\mathcal{S}(\theta) =\mathcal{L}(f_{\theta }(\mathcal{S})),
\end{equation}
where $\mathcal{L}(\cdot)$ refers to the loss function (\eg, the cross-entropy loss $\mathcal{L}_{CE}$ \wrt Eq.~\eqref{eq:ce}). Then, we make a temporary copy for the original parameters $\theta$ and update the intermediate parameters for fast weights as follows:
\begin{equation}
\label{eq:spt_update}
    \theta'_i = \theta-\alpha \nabla_{\theta} \mathcal{L}_{\mathcal{T}_i}^{\mathcal{S}}(\theta),
\end{equation}
where $\alpha$ denotes the learning rate for \textit{meta training}. Intuitively, this step simulates a direct update to train the learner to obtain class-specific knowledge of the support data.

\noindent {\bf Meta testing.}
After meta training on the support batch, we then scheme a virtual testing process, leveraging the query batch $\mathcal{Q}=\{V_i, T_i\}^B$, where $\mathcal{S}\cap \mathcal{Q}=\emptyset$,  to evaluate the generalization performance of the base learner $f_{\theta_i'}$. Formally, we measure the known-to-open performance on $\mathcal{T}_i$ by calculating the class-specific loss \wrt the query data as:
\begin{equation}
\label{eq:qry}
    \mathcal{L}_{\mathcal{T}_i}^\mathcal{Q}(\theta'_i) =\mathcal{L}(f_{\theta_i' }(\mathcal{Q})).
\end{equation}
Here, the formulation closely relates to the standard fine-tuning process, which aims to obtain decent class-specific performance for all training batches. Differently, this step merely evaluates the intermediary base learner for its known-to-open generalizability on each task, due to the original parameters $\theta$ remaining immune to the task-specific updates.
Hence, it can be used to provide feedback for the learner on \textit{what video-specific knowledge should be learned to derive the robust generalization} against inherent known and static biases in the following meta-optimization.

\noindent {\bf Meta-optimization.}
As mentioned above, the intuition behind our approach is that the virtual evaluation during meta testing can provide useful feedback to encourage the learning of more robust representations for fast known-to-open generalization after \textit{meta training} on the support data (\ie, $\theta'_i \leftarrow \theta$).
Note that original MAML approaches focus on optimizing parameters for a strong initialization, enabling quick adaptation to new tasks with minimal gradient updates. Conversely, open-vocabulary recognition requires zero-shot capabilities, where no further adaptation can be applied for new tasks. Therefore, class-specific knowledge should be strengthened in terms of global optimization.
To this end, within the outer loop, the parameters of the learner are optimized to minimize the class-specific errors for the support data and the adaptation cost for the query data simultaneously. 
The combination of both Eq.~\eqref{eq:spt} and Eq.~\eqref{eq:qry} is used to carry out the outer loop update, thus the objective for meta-optimization can be defined as:
\begin{equation}
\label{eq:maml*}
\begin{aligned}
\min _{\theta}\mathcal{L}_{\mathcal{T}_i}(\theta )=& \min _{\theta}\left( \mathcal{L}_{\mathcal{T}_i}^\mathcal{S} (\theta) +\mathcal{L}_{\mathcal{T}_i}^\mathcal{Q}(\theta_i') \right)  \\
= & \min _{\theta} \left(\mathcal{L}_{\mathcal{T}_i}^\mathcal{S} (\theta)+\mathcal{L}_{\mathcal{T}_i}^\mathcal{Q}(\theta-\alpha \nabla_{\theta} \mathcal{L}_{\mathcal{T}_i}^{\mathcal{S}}(\theta))\right).
\end{aligned}
\end{equation}
Here, the first term refers to the class-specific knowledge learned on the support batch, while the second term provides the known-to-open generalization feedback based on $\theta_i'$ towards robust representation learning \wrt the task $\mathcal{T}_i$.
The optimizing process of the parameter $\theta$ can be given by:
\begin{equation}
\label{eq:maml**}
    \theta \leftarrow \theta - \beta \nabla_{\theta} \sum _{i = 1}^{N}\left( \mathcal{L}_{\mathcal{T} _i}^\mathcal{S} (\theta) + \mathcal{L}_{\mathcal{T} _i}^\mathcal{Q} \left ( \theta - \alpha \nabla_{\theta} \mathcal{L}_{\mathcal{T} _i}^\mathcal{S} (\theta) \right) \right),
\end{equation}
where $N$ is the batch size of the task for meta-optimization.
Since the MAML meta-gradient update needs to differentiate through the optimization process (\ie, a gradient by a gradient), it’s not an ideal solution where we need to optimize a large number of tasks during the training phase. 
Therefore, we opt for the one-step update approximation by dropping the backward pass of $\theta \gets \theta'_i$ as:
\begin{equation}
\label{eq:FOMAML}
    \theta \leftarrow \theta - \beta  \sum _{i=1}^{N}\left (\nabla_{\theta}\mathcal{L}_{\mathcal{T} _i}^\mathcal{S} (\theta) +\delta \nabla_{\theta'_i} \mathcal{L}_{\mathcal{T}_i}^{\mathcal{Q}}(\theta'_i)\right),
\end{equation}
where $\beta$ and $\delta$ are the learning rates for meta-optimization.
With the genuine update of the learner in Eq.~\eqref{eq:FOMAML} without CLIP regularization, we can optimize a parallel or batch version that evaluates on $N$ known-to-open tasks of different class distributions (\ie, class-specific knowledge), which encourages to learn more generalizable features against known and static biases.

\begin{table*}[!ht]
\centering
\caption{Performance comparison (Top1-Acc (\%)) with the CLIP-adapted methods using ViT-B/16 under the in-context base-to-novel setting. We also report the harmonic mean (HM) of base and novel recognition accuracy. The \textbf{best} and the \underline{second-best} results are highlighted. $\ast$ and $\dagger$ denote the results reproduced with our implementation using frozen text learners.}
\label{tab:B2N}
\resizebox{0.9\textwidth}{!}{
\begin{tabular}{lccccccccccccc}
\toprule
\multirow{2}{*}{Method} &
\multirow{2}{*}{Venue} &
  \multicolumn{3}{c}{K400} &
  \multicolumn{3}{c}{HMDB} &
  \multicolumn{3}{c}{UCF} &
  \multicolumn{3}{c}{SSv2} \\ \cmidrule(lr){3-5} \cmidrule{6-8} \cmidrule{9-11} \cmidrule{12-14} 
 & &
  Base &
  Novel &
  \multicolumn{1}{c|}{HM} &
  Base &
  Novel &
  \multicolumn{1}{c|}{HM} &
  Base &
  Novel &
  \multicolumn{1}{c|}{HM} &
  Base &
  Novel &
  HM \\ \cmidrule{1-14}
Frozen CLIP~\cite{radford2021learning} & ICML'21 &
  62.3 &
  53.4 &
  \multicolumn{1}{c|}{57.5} &
  53.3 &
  46.8 &
  \multicolumn{1}{c|}{49.8} &
  78.5 &
  63.6 &
  \multicolumn{1}{c|}{70.3} &
  4.9 &
  5.3 &
  5.1 \\
ActionCLIP~\cite{wang2021actionclip} & arXiv'21 &
  61.0 &
  46.2 &
  \multicolumn{1}{c|}{52.6} &
  69.1 &
  37.3 &
  \multicolumn{1}{c|}{48.5} &
  90.1 &
  58.1 &
  \multicolumn{1}{c|}{70.7} &
  13.3 &
  10.1 &
  11.5 \\
X-CLIP~\cite{ni2022expanding} & ECCV'22 &
  74.1 &
  56.4 &
  \multicolumn{1}{c|}{64.0} &
  69.4 &
  45.5 &
  \multicolumn{1}{c|}{55.0} &
  89.9 &
  58.9 &
  \multicolumn{1}{c|}{71.2} &
  8.5 &
  6.6 &
  7.4 \\
VPT~\cite{ju2022prompting} & ECCV'22 &
  69.7 &
  37.6 &
  \multicolumn{1}{c|}{48.8} &
  46.2 &
  16.0 &
  \multicolumn{1}{c|}{23.8} &
  90.5 &
  40.4 &
  \multicolumn{1}{c|}{55.8} &
  8.3 &
  5.3 &
  6.4 \\
ST-Adapter~\cite{pan2022st} & NeurIPS'22 &
  74.6 &
  62.0 &
  \multicolumn{1}{c|}{67.3} &
  65.3 &
  48.9 &
  \multicolumn{1}{c|}{55.9} &
  85.5 &
  76.8 &
  \multicolumn{1}{c|}{80.9} &
  9.3 &
  8.4 &
  8.8 \\
ViFi-CLIP~\cite{rasheed2023fine} & CVPR'23 &
  \underline{76.4} &
  61.1 &
  \multicolumn{1}{c|}{67.9} &
  \textbf{73.8} &
  \underline{53.3} &
  \multicolumn{1}{c|}{\underline{61.9}} &
  92.9 &
  67.7 &
  \multicolumn{1}{c|}{78.3} &
  \underline{16.2} &
  \underline{12.1} &
  \underline{13.9} \\
Open-VCLIP $\ast$~\cite{weng2023open} & ICML'23 &
  76.3 &
  \underline{62.3} &
  \multicolumn{1}{c|}{\underline{68.6}} &
  70.2 &
  50.2 &
  \multicolumn{1}{c|}{58.5} &
  \underline{94.6} &
  \underline{77.2} &
  \multicolumn{1}{c|}{\underline{85.0}} &
  15.9 &
  10.8 &
  12.9 \\
FROSTER $\dagger$~\cite{huang2024froster}& ICLR'24 &
  76.0 &
  61.9 &
  \multicolumn{1}{c|}{68.3} &
  70.0 &
  49.9 &
  \multicolumn{1}{c|}{58.3} &
  94.3 &
  76.9 &
  \multicolumn{1}{c|}{84.7} &
  15.5 &
  10.3 &
  12.4 \\
\rowcolor{red!10} 
\textbf{Open-MeDe} & &
  \textbf{77.2} &
  \textbf{63.8} &
  \multicolumn{1}{c|}{\textbf{69.9}} &
  \underline{73.6} &
  \textbf{56.4} &
  \multicolumn{1}{c|}{\textbf{63.9}} &
  \textbf{94.9} &
  \textbf{78.5} &
  \multicolumn{1}{c|}{\textbf{85.9}} &
  \textbf{17.1} &
  \textbf{12.3} &
  \textbf{14.3} \\ \bottomrule
\end{tabular}
}
\end{table*}
\vspace{-0.3em}
\subsubsection{Gaussian self-ensemble stabilization}
Typically, training the video learner for longer iterations to gain specialization on the supervised tasks comes with the risk of diminished plasticity and generalizability. 
Model patching~\cite{ilharco2022patching,wortsman2022robust,shu2023clipood,weng2023open} of weight ensembling has been shown to improve both the performance and generalization.
Given that the fine-tuning videos are limited in class-specific knowledge, while the open-vocabulary tasks are unconstrained, the static generalizable flexibility derived from large-scale I-VL pre-training should be scrupulously exploited to enhance the adaptation of the video learner while minimizing the impact of static bias.
Therefore, we further incorporate self-ensemble stabilization tailored to the video learner over its optimization trajectory, which utilizes the knowledge from previous training iterations for a generalizable solution.
In a fine-tuning procedure of $R$ epochs with $l$ step length for each, the learner's optimization trajectory is represented by $\{\theta_t\}_{t=1}^R$, and $\theta_0$ is the pre-trained weights. The self-ensemble averages the weights of the learner as:
\begin{equation}
    \theta_{\texttt{WA}}=(1-\sum_{t=1}^{R}\alpha_t)\cdot\theta_0 + \sum_{t=1}^{R}{\alpha_t}\cdot \theta_t,
\end{equation}
where $\alpha_t\in[0,1]$ specifies the weights contributed by the parameters at $t$-th epoch.
Intuitively, during the early fine-tuning epochs (\ie, at a smaller epoch $t$), the video learner lacks the maturity to effectively capture video-specific knowledge while still retaining substantial static-related orientation from large-scale pre-training, which introduces vulnerable information for temporal understanding. 
Conversely, the parameters at the last few epochs (\ie, at a larger epoch $t$) have integrated more video-specific knowledge, highly featuring the supervised downstream task distribution, whereas the plasticity of the unconstrained zero-shot capability is not guaranteed.
As both sides degrade the final open-vocabulary generalizability, we aim to weaken the contribution of the parameters near the initial and terminal epochs by employing a distribution prior, resulting in a generic optima for robust generalization.
\begin{algorithm}[!t]
\small
   \caption{Training Procedure}
   \label{alg:training}
   \SetKwInOut{Require}{Require}
   \KwIn{Training set $D_{tr}=\{V_i,T_i\}^M$, Video learner $f_\theta$.}  
   \Require{GWA Params $\theta_{\texttt{GWA}}$ update at each epoch with $l$ step length. CLIP Params $\theta_{\texttt{CLIP}}$. Batch size of training samples $B$. Learning rate $\alpha,\beta,\delta$.}
   \KwOut{The final GWA learner $f_{\theta_{\texttt{GWA}}}$.}
   \BlankLine
   Initialize $\theta, \theta_{\texttt{GWA}} \gets \theta_{\texttt{CLIP}}; \text{Step}=0; t=0$\\
   \While{not coverged}{
   $\text{Step} \gets \text{Step}+1$ \\
   Construct batch of tasks $\mathcal{T}_i=\{\mathcal{S},\mathcal{Q}\}$ by sampling $\mathcal{S},\mathcal{Q} \gets \{V_a,T_a\}^B, \{V_b,T_b\}^B \subseteq D_{tr}$\\
   \ForAll {$\mathcal{T}_i$}{
   \textcolor{iccvblue}{\tcp{meta training}}
   Evaluate $\nabla_\theta \mathcal{L}_{\mathcal{T}_i}^{\mathcal{S}}(\theta)$ \wrt Eq.~\eqref{eq:spt} \\
   Compute adapted parameters with gradient decent: $\theta'_i = \theta-\alpha \nabla_{\theta} \mathcal{L}_{\mathcal{T}_i}^{\mathcal{S}}(\theta)$ \wrt Eq.~\eqref{eq:spt_update}\\
   }
   \textcolor{iccvblue}{\tcp{meta testing}}
   Evaluate $\nabla_{\theta'_i} \mathcal{L}_{\mathcal{T}_i}^{\mathcal{Q}}(\theta'_i)$ \wrt Eq.~\eqref{eq:qry}\\
   \textcolor{iccvblue}{\tcp{meta-optimization}}
   Update $\theta$ \wrt Eq.~\eqref{eq:FOMAML} \\
   \textcolor{iccvblue}{\tcp{Gaussian Weight Average}}
   \If{$\text{mod}(\text{Step}, l)==0$}{
   $t \gets t+1; \theta_t \gets \theta$ \\
   Update $\theta_{\texttt{GWA}}$ \wrt Eq.~\eqref{eq:GWA}\\
   }
   }
\end{algorithm}
   
Driven by~\cite{khattak2023self} in prompt learning, we perform Gaussian Weight Average (GWA) based on model patching, as shown in~\cref{fig:overview}(b), which assigns the parameters with lower weights at initial epochs, higher weights at middle epochs, and relatively lower weights at final epochs.
Given a  Gaussian distribution $w_t\sim \mathcal{N}(\mu,\sigma^2)$ defined over the epochs, we sample the weight values for the parameters $\theta_t$ as its corresponding probability in the distribution as:
\begin{equation}
    w_t=\frac{1}{\sqrt{2 \pi} \sigma} e^{-\frac{(t-\mu)^{2}}{2 \sigma^{2}}}, t=1,\dots,R.
\end{equation}
Here, we exclude the integration of CLIP weights $\theta_0$ for the purpose of static debiasing. $\mu$ and $\sigma^2$ are hyper-parameters for the distribution, and in practice, we determine the value of $\mu$ according to the epoch number. Then, we perform normalization towards the weights of total epochs \ie, $\alpha_t = \frac{w_t}{\sum_{i=1}^{R}w_i}$. 
We also formulate GWA as a moving average to avoid increasing the storage cost of saving multiple snapshots of the parameters by updating the average of current learner $\theta_t$ on the fly (\ie, at epoch $t$) as:
\begin{equation}
\label{eq:GWA}
    \theta_{\texttt{GWA}}\gets \frac{\sum_{i=1}^{t-1} w_{i}}{\sum_{i=1}^{t} w_{i}} \cdot \theta_{\texttt{GWA}}+\frac{w_{t}}{\sum_{i=1}^{t} w_{i}} \cdot \theta_{t} .
\end{equation}

\subsection{Algorithm overview}
We present the overall training procedure of the proposed model-agnostic Open-MeDe in~\cref{alg:training}. The video learner is fine-tuned based on training videos based on our cross-batch meta-optimization scheme cost-effectively.
And the Gaussian self-ensemble stabilization is performed on the video learner via our GWA for robust generalization under open-vocabulary settings.

\section{Experiments}
\label{sec:experiments}

\subsection{Experimental Setup}

\noindent {\bf Datasets.}
We explore two distinct types of open-vocabulary action recognition evaluation in this work: \textit{in-context} and \textit{out-of-context} settings.
For in-context scenarios, we conduct experiments following the common practice in the literature~\cite{rasheed2023fine,weng2023open,huang2024froster,rasheed2023fine} on the Kinetics-400 (K400)~\cite{k400}, UCF-101 (UCF)~\cite{UCF101}, HMDB-51 (HMDB)~\cite{HMDB51}, Something-Something V2 (SSv2)~\cite{ssv2} and Kinectics-600 (K600)~\cite{k600} datasets under widely-used evaluation protocols: \textit{cross-dataset} and \textit{base-to-novel} evaluation. For more challenging out-of-context scenarios, we newly conduct general cross-dataset evaluations using K400 dataset as the training set and testing on the synthetic UCF-SCUBA~\cite{li2023mitigating} and UCF-HAT~\cite{chung2022enabling,bae2023devias} benchmarks. 

\noindent {\bf Implementation details.}
Generally, we use the official CLIP ViT-B/16 backbone for all experiments, and our video learner is the adaptation of the CLIP model follows~\cite{weng2023open}, unless stated otherwise.
During our meta-optimization process, we construct a batch of 4 tasks, each task contains 8 support and query samples from the training set. The learning rates of inner and outer loops for support batches \ie, $\alpha$, and $\beta$, are synchronized with the initial value of $3.33\times 10^{-6}$ and decay to $3.33\time 10^{-8}$ utilizing the AdamW~\cite{loshchilov2017decoupled} optimizer following a cosine decay scheduler, while the hyperparameter $\delta$ for query batches is set to $1.67\times 10^{-3}$. 
For cross-dataset evaluation, we warm up the training on the K400 dataset for the first 2 epochs and further fine-tune the video learner for 20 epochs. For base-to-novel evaluation, we train the learner for 12 epochs with the first two warm-up epochs on training data. 
During inference, we use 3 temporal and 1 spatial views per video and linearly aggregate the recognition results. 
See Appendix for experimental details.

\subsection{Comparison with state-of-the-art methods}
We compare our framework with the state-of-the-art open-vocabulary action recognition methods on the following commonly used \textit{in-context} and newly proposed \textit{out-of-context} evaluation protocols.

\noindent {\bf In-context base-to-novel generalization.} 
In~\cref{tab:B2N}, we compare the proposed framework with other CLIP-based methods under the popular in-context base-to-novel setting. 
All methods are initially learned on the frequently occurring base classes and evaluated on both base and novel classes, where the novel classes represent a realm of previously uncounted scenarios.
From the results, we can summarize the observations: (1) Most of the methods show reasonable improvements from the frozen CLIP~\cite{radford2021learning}, except for ActionCLIP~\cite{wang2021actionclip}, X-CLIP~\cite{ni2022expanding} and VPT~\cite{ju2022prompting} suffering inferior performances especially on the novel sets of K400, HMDB and UCF, indicating the strong generalization of CLIP and the potential overfitting of these adapted video learners toward the training samples.
(2) Our framework experiences noticeable gains in novel class performance and consistent achievements on all four datasets, spanning spatially dense and temporally focused scenarios, which validates the effectiveness of enhancing generalization and static debiasing for both known and open classes.

\begin{table}[!t]
\small
\centering
\caption{Comparison with the previous methods under the in-context cross-dataset setting. The results are top-1 accuracies (\%) with mean and standard deviation on the evaluation across three validation splits within each dataset. $\ast$ and $\dagger$ denote our re-implementation with frozen text learners.}
\label{tab:cross-dataset}
\resizebox{\columnwidth}{!}{
\begin{tabular}{lcccc}
\toprule
Method &  \multicolumn{1}{c}{Venue}   & \multicolumn{1}{c}{UCF} & \multicolumn{1}{c}{HMDB} & \multicolumn{1}{c}{K600} \\ \cmidrule{1-5}
Frozen CLIP~\cite{radford2021learning} & ICML'21  & 73.8$\pm$0.6                & 47.9$\pm$0.5                 & 68.1$\pm$1.1                 \\
ActionCLIP~\cite{wang2021actionclip} & arXiv'21   & 77.5$\pm$0.8                & 48.2$\pm$1.5                 & 62.5$\pm$1.2                 \\
X-CLIP~\cite{ni2022expanding}&ECCV'22 & 72.0$\pm$2.3                & 44.6$\pm$5.2                 & 65.2$\pm$0.4                 \\
VPT~\cite{ju2022prompting}&ECCV'22          & 69.3$\pm$4.2                & 44.3$\pm$2.2                 & 55.8$\pm$0.7                 \\
ST-Adapter~\cite{pan2022st}&NeurIPS'22   & 77.6$\pm$0.7                & 51.1$\pm$0.6                 & 60.2$\pm$1.8                 \\
Vita-CLIP~\cite{wasim2023vita}&CVPR'23    & 75.0$\pm$0.6                & 48.6$\pm$0.6                 & 67.4$\pm$0.5                 \\
MAXI~\cite{lin2023match}&ICCV'23         & 78.2$\pm$0.7                & 52.3$\pm$0.6                 & 71.5$\pm$0.8                 \\
Open-VCLIP $\ast$~\cite{weng2023open} &ICML'23  & \underline{83.3$\pm$1.4}                & \underline{53.8$\pm$1.5}                 & \underline{73.0$\pm$0.8}                 \\
ViLT-CLIP~\cite{wang2024vilt}&AAAI'24    & 73.6$\pm$1.1                & 45.3$\pm$0.9                 & -                        \\
FROSTER $\dagger$~\cite{huang2024froster}&ICLR'24    & 82.9$\pm$0.6                   & 53.4$\pm$1.2                 & 71.1$\pm$0.8                 \\
VicTR~\cite{kahatapitiya2024victr}&CVPR'24        & 72.4$\pm$0.3                & 51.0$\pm$1.3                 & -                        \\
ALT~\cite{chen2024align}& CVPR'24          & 79.4$\pm$0.9                & 52.9$\pm$1.0                 & 72.7$\pm$0.6                 \\
\rowcolor{red!10}
\textbf{Open-MeDe } &       & \textbf{83.7$\pm$1.3}                    & \textbf{54.6$\pm$1.1}                 & \textbf{73.7$\pm$0.9}   \\
\bottomrule
\end{tabular}
}
\end{table} 
\noindent {\bf In-context cross-dataset generalization.}
In~\cref{tab:cross-dataset}, we present the compared results under in-context cross-dataset zero-shot evaluations, where all learners undergo further fine-tuning on K400 training set and are tested directly on downstream cross-datasets \ie, UCF, HMDB and K600.
Similar findings can be noticed from the results as base-to-novel evaluations that frozen CLIP outperforms several adapted learners, especially on the most generalizability demanding benchmark, \ie, K600, further demonstrating the generalization degradation of overfitting within these methods.
Remarkably, our framework based on meta-learning consistently surpasses state-of-the-art approaches on all three benchmarks, demonstrating its superior effectiveness and enhanced generalizability.

\begin{table}[!t]
\caption{Performance comparison (Top-1 / Top-5 Acc. (\%)) on UCF dataset. We evaluate both in-context and out-of-context recognition (marked with $\star$) performances. We also report the harmonic mean (HM) of the results. $\ast$ and $\dagger$ indicate our implementation with frozen text learners.}
\label{tab:ooc}
\resizebox{\columnwidth}{!}{
\centering
\begin{tabular}{lcccc}
\toprule
Method     & UCF         & UCF-SCUBA $\star$   & UCF-HAT $\star$     & HM          \\ \cmidrule{1-5}
X-CLIP     & 74.5 / 95.4   & 24.6 / 43.3   & 56.8 / 78.1   & 20.3 / 64.7   \\
Open-VCLIP $\ast$ & \underline{83.5 / 96.9} & \underline{28.9 / 48.0} & \underline{59.6 / 79.5} & \underline{47.4 / 68.6} \\
FROSTER $\dagger$    & 82.9 / 96.4 & 25.2 / 43.2 & 58.6 / 78.9 &  43.6 / 64.9 \\
\rowcolor{red!10}
\textbf{Ours} & \textbf{83.9 / 96.9} & \textbf{33.5 / 52.7} & \textbf{64.5 / 82.3} & \textbf{52.4 / 72.4} \\ \bottomrule
\end{tabular}
}
\end{table}
\noindent {\bf Out-of-context cross-dataset generalization.}
In~\cref{tab:ooc}, we further compare our method with the previous state of the arts under more challenging out-of-context cross-dataset evaluations on SCUBA and HAT benchmarks of the UCF dataset.
It can be noticed that: 
(1) Integrating with CLIP regularization, both Open-VCLIP~\cite{weng2023open} and FROSTER~\cite{huang2024froster} achieve promising improvements compared with X-CLIP under original UCF in-context scenarios.
(2) However, the compared methods suffer from severely limited generalization when encountering out-of-context scenarios due to the static bias within these video learners.
(3) Our method significantly outperforms partially fine-tuned X-CLIP and CLIP regularization methods on various out-of-context scenarios. We outperform the second-best competitor by $4.6\%$ on UCF-SCUBA and $4.9\%$ on UCF-HAT, with the highest HM striking an impressive balancing on cross-dataset generalization for in-context and out-of-context scenarios. We attribute the superiority of our video learner to the natural know-to-open generalizing and image-to-video debiasing via the newly proposed meta-optimization and self-ensemble independent from CLIP's persistent interference of static biases for robust and generic generalizability.

\subsection{Ablation Studies}
\begin{table}[!t]
\centering
\caption{In-context cross-dataset comparison (Top-1 Acc. (\%)) when integrating our Open-MeDe with different video learners.}
\label{tab:applicability}
\resizebox{\columnwidth}{!}{
\begin{tabular}{l|lccc}
\toprule
Adaptation                            & Method       & UCF      & HMDB     & K600     \\ \cmidrule{1-5}
\multirow{3}{*}{Adapter-based}        & ST-Adapter~\cite{pan2022st}   & 77.6$\pm$0.7 & 51.1$\pm$0.6 & 60.2$\pm$1.8 \\
                                      &\cellcolor{gray!20} $+$ \textbf{Ours}       &\cellcolor{gray!20}  \textbf{78.9$\pm$1.1}        &\cellcolor{gray!20} \textbf{52.0$\pm$1.1}     &\cellcolor{gray!20} \textbf{72.7$\pm$0.8}          \\
                                      & $\Delta$ Gains        & \textcolor{blue!60}{$+$ \textbf{1.3}}         & \textcolor{blue!60}{$+$ \textbf{0.9}}      & \textcolor{blue!60}{$+$ \textbf{12.5}}       \\ \cmidrule{1-5}
\multirow{3}{*}{Prompt-based}         & Vita-CLIP~\cite{wasim2023vita}    & 75.0$\pm$0.6 & 48.6$\pm$0.6 & 67.4$\pm$0.5 \\
                                      & \cellcolor{gray!20}$+$ \textbf{Ours}       & \cellcolor{gray!20} \textbf{77.9$\pm$0.8}        & \cellcolor{gray!20} \textbf{50.7$\pm$1.3}      & \cellcolor{gray!20}\textbf{71.5$\pm$0.9}        \\ 
                                      & $\Delta$ Gains       & \textcolor{blue!60}{$+$ \textbf{2.9}}         & \textcolor{blue!60}{$+$ \textbf{2.1}}      & \textcolor{blue!60}{$+$ \textbf{4.1}}       \\ \cmidrule{1-5}
\multirow{3}{*}{Partially-tuned} & X-CLIP~\cite{ni2022expanding}        & 72.0$\pm$2.3 & 44.6$\pm$5.2 & 65.2$\pm$0.4 \\
                                      &\cellcolor{gray!20} $+$ \textbf{Ours} &\cellcolor{gray!20} \textbf{79.3$\pm$1.3}         &\cellcolor{gray!20}  \textbf{52.3$\pm$1.5}      &\cellcolor{gray!20} \textbf{72.9$\pm$1.1}      \\ 
                                      & $\Delta$ Gains       & \textcolor{blue!60}{$+$ \textbf{7.3}}         & \textcolor{blue!60}{$+$ \textbf{7.7}}      & \textcolor{blue!60}{$+$ \textbf{7.7}}         \\ \cmidrule{1-5}
\multirow{3}{*}{Fully-tuned}          & VCLIP~\cite{weng2023open}        & 78.5$\pm$1.0 & 50.3$\pm$0.8 & 65.9$\pm$1.0 \\
                                      &\cellcolor{gray!20} $+$ \textbf{Ours}       & \cellcolor{gray!20}\textbf{83.7$\pm$1.3}     & \cellcolor{gray!20}\textbf{54.6$\pm$1.1}        & \cellcolor{gray!20}\textbf{73.7$\pm$0.9 }       \\ 
                                      & $\Delta$ Gains      & \textcolor{blue!60}{$+$ \textbf{5.2}}           & \textcolor{blue!60}{$+$ \textbf{4.3}}           & \textcolor{blue!60}{$+$ \textbf{7.8}}       \\ \bottomrule
\end{tabular}
}
\end{table}
\noindent {\bf Applicability with different video learners.}
In~\cref{tab:applicability}, we adopt other video learners (with the frozen text encoder) from adapter-based ST-Adapter~\cite{pan2022st}, prompt-based Vita-CLIP~\cite{wasim2023vita}, partially fine-tuned X-CLIP~\cite{ni2022expanding} and fully fine-tuned VCLIP~\cite{weng2023open} to validate the effectiveness of our model-agnostic framework.
We find that: (1) All CLIP-adapted video learners integrating with our method achieve consistent improvements on in-context cross-dataset evaluations, highlighting its broad and flexible applicability.
(2) Our approach generally exhibits more improvements for partially and fully fine-tuned methods than PEFT learners, suggesting the importance of sufficient fitting capacity (\ie, learnable parameters) for video learners to attain video-specific generalizability.

\begin{figure}[!t]
    \centering
    \includegraphics[width=\columnwidth]{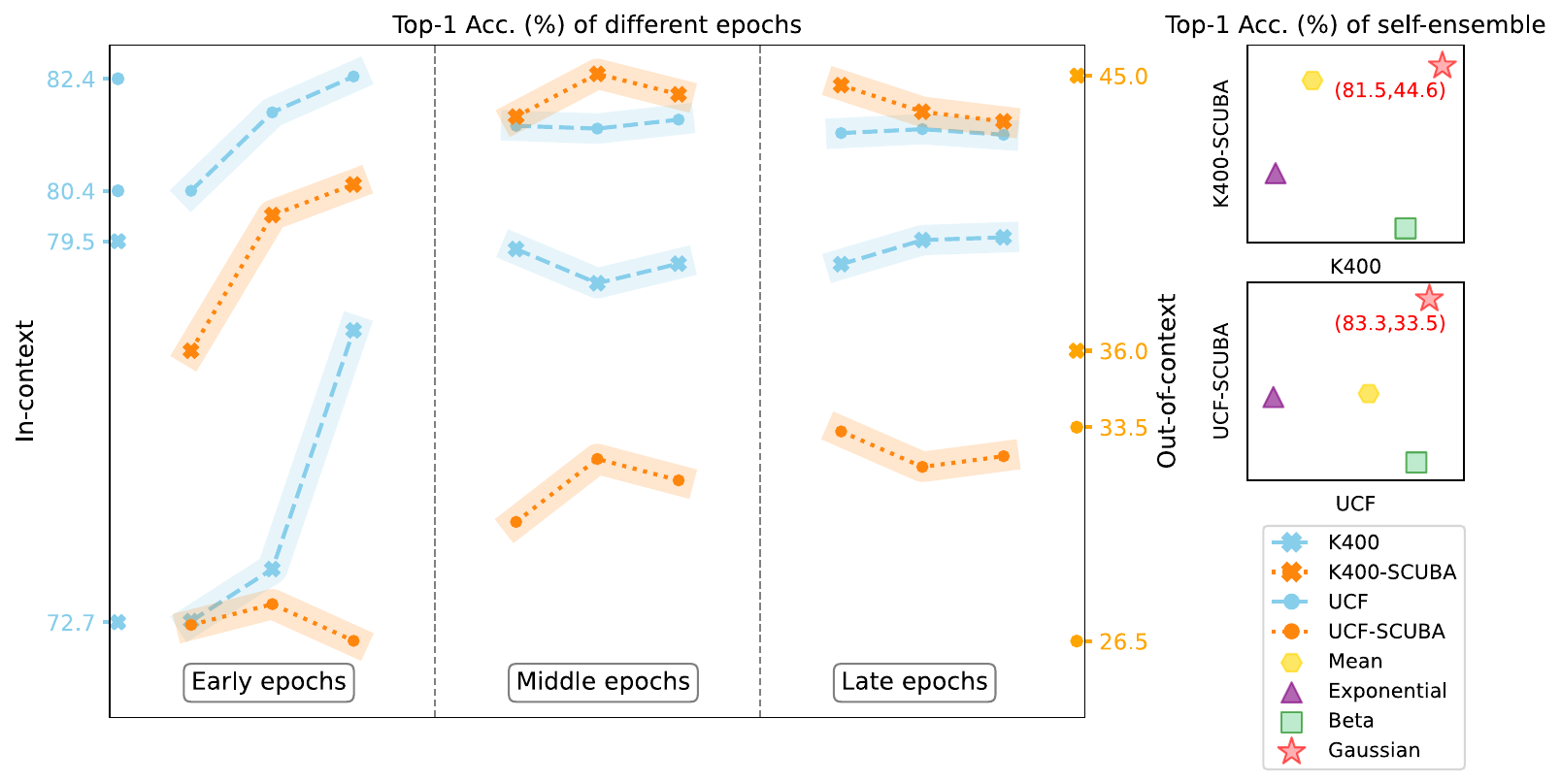}
    \caption{Performance comparison at different epochs \vs various weight self-ensemble strategies. We train the video learner on K400 and test on the in-context UCF, K400, and out-of-context K400-SCUBA and UCF-SCUBA benchmarks. Points on the curves represent epochs of $[2,4,6]$, $[10,12,14]$ and $[18,20,22]$ from left to right, respectively.
 }
    \label{fig:k400-ucf}
\end{figure}
\begin{table}[!t]
\caption{We compare the performances of different optimization schemes under various settings. IC: in-context evaluations, OC: SCUBA~\cite{li2023mitigating} out-of-context evaluations, HM: harmonic mean. RFD: Residual Feature Distillation, IWR: Interpolated Weight Regularization, Meta Unseen: MAML for meta seen to unseen, Meta Cross-batch: our cross-batch meta-optimization.}
\label{tab:optim}
\resizebox{\columnwidth}{!}{
\begin{tabular}{l|l|ccc|ccc}
\toprule
\multirow{2}{*}{Optimization} & \multirow{2}{*}{Method} & \multicolumn{3}{c|}{K400 (closed-set)} & \multicolumn{3}{c}{UCF (zero-shot)} \\
 &  & IC & OC & HM & IC & OC & \multicolumn{1}{l}{HM} \\ \cmidrule{1-8}
Plain & (a) VCLIP~\cite{weng2023open} & 80.1 & \underline{42.4} & \underline{55.4} & 78.5 & 28.3 & 41.6 \\ \cmidrule{1-8}
\multirow{2}{*}{CLIP Reg.} 
 & (b) + RFD~\cite{huang2024froster} & 79.9 & 41.5 & 54.6 & 82.5 & 25.2 & 38.9 \\ 
& (c) + IWR~\cite{weng2023open} & \underline{80.5} & 40.3 & 53.7 & 82.9 & 28.9 & 42.9 \\
 \cmidrule{1-8}
\multirow{2}{*}{Meta learning} & (d) + Meta Unseen~\cite{verma2020meta} & 79.5 & 41.7 & 54.7 & \underline{83.2} & \underline{31.8} & \underline{46.0} \\
 & \cellcolor{gray!20}(e) + Meta Cross-batch & \cellcolor{gray!20}\textbf{81.5} & \cellcolor{gray!20}\textbf{46.6} & \cellcolor{gray!20}\textbf{59.3} & \cellcolor{gray!20}\textbf{83.9} & \cellcolor{gray!20}\textbf{33.5} & \cellcolor{gray!20}\textbf{47.9} \\ \bottomrule
\end{tabular}
}
\vspace{-0.2cm}
\end{table}
\noindent {\bf Effect of cross-batch meta-optimization.}
In~\cref{tab:optim}, we conduct experiments to verify the effect of our cross-batch meta-optimization scheme. The compared strategies and analyses are as follows: (a) Consider VCLIP with standard fine-tuning objectives as a baseline of the plain learner. (b) When adopting RFD to VCLIP, the K400 closed-set performance experiences a slight decline for both IC and OC scenarios, while cross-dataset in-context generalization improves, with gains of $+4.5\%$ on UCF-IC, whereas it severely impairs generalization for UCF-OC ($-3.1\%$). (c) Similar results are observed when integrating IWR regularization with VCLIP. (d) For the previous meta unseen optimization method for zero-shot learning, all three accuracies under UCF cross-dataset evaluation increase, where K400 evaluations challenge its closed-set generalizations, indicating the potential overfitting to meta unseen tasks.
(e) Notably, our cross-batch meta-optimization scheme ((a)$\rightarrow$(e)) enhances all closed-set and zero-shot performance on harmonic mean with gains of $+3.9\%$ and $+6.3\%$, respectively. This showcases the superiority of our scheme for enhancing know-to-open generalizing and image-to-video debiasing, which establishes a promising balance for robust generalization capabilities.

\noindent {\bf Effect of weight self-ensemble.}
In~\cref{fig:k400-ucf}, we investigate the trend of generalization performance during K400 training and the efficacy of weight self-ensemble stabilization using various strategies. 
In particular, the curves illustrate the performance within the video learner's optimization trajectory at different epochs, where the $x$-axis and $y$-axis display the different stages of training epochs and various generalization evaluation protocols, respectively.
It is noticeable that the overall performance has experienced trends of significant enhancement on both closed-set and zero-shot generalization while quickly leading to drops in zero-shot performance at the tail of the fine-tuning phase, suggesting the plasticity degradation that highly features supervised task-specific distributions on the downstream dataset.
The results show that weight ensembling methods improve both specialty and generalizability, with our Gaussian self-ensemble excelling significantly, strongly suggesting it as a better choice for robust generalization.

\begin{figure}[t]
    \centering
    \includegraphics[width=0.8\columnwidth]{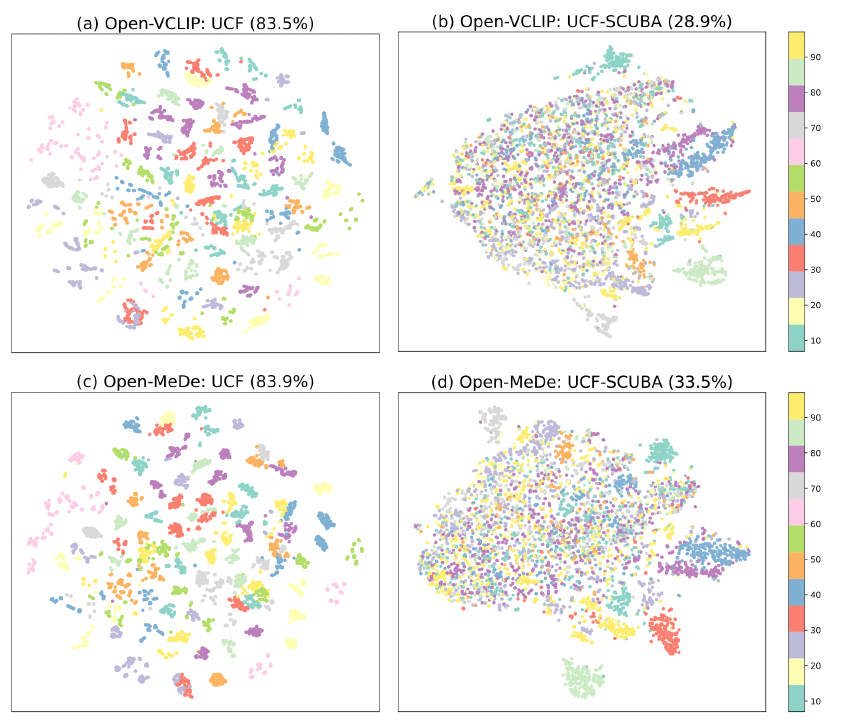}
    \caption{t-SNE~\cite{van2008visualizing} visualization of the predictions from Open-VCLIP and our Open-MeDe on UCF and UCF-SCUBA.}
    \label{fig:visual}
    \vspace{-0.5cm}
\end{figure}
\subsection{Visualizations} 
\cref{fig:visual} compares the t-SNE visualizations of Open-VCLIP and our framework for in-context and out-of-context UCF predictions. Note that our predictions for videos within the same category are more concentrated, with reduced confusion between different categories, compared to Open-VCLIP. This suggests that the proposed framework effectively learns temporal information, mitigating known and static biases while demonstrating robust generalizability. However, there remains considerable room for improvement in out-of-context scenarios for video-adapted learners.

\section{Conclusion}
\label{sec:conclusion}
We introduce Open-MeDe, a novel meta-learning framework for open-vocabulary action recognition. 
It adopts a cross-batch meta-optimization, which encourages the video learner to attain generalizable knowledge counteracting inherent known and static biases for effective known-to-open generalizing and image-to-video debiasing.
It also incorporates Gaussian Weight Average to achieve generic optima for robust generalization.
Extensive evaluations in both in-context and out-of-context open-vocabulary scenarios validate the applicability and superiority of our framework.
\section*{Acknowledgments}
This research is supported by the National Natural Science Foundation of China (No. 62376217, 62273347, and 62301434), and the Young Elite Scientists Sponsorship Program by CAST (No. 2023QNRC001).

{
    \small
    \bibliographystyle{ieeenat_fullname}
    \bibliography{main}
}

\setcounter{page}{1}
\appendix
\maketitlesupplementary
This supplementary material provides additional details and further experiments to complement the main paper. The content is organized as follows:
\begin{enumerate}[label=\Alph*.]
    \item Additional Experimental Details (\appenref{experimental details})
    \item Additional Experimental Results (\appenref{ad-results})
    \item Discussions (\appenref{discussion})
    \item Broader Impacts and Limitations (\appenref{limitation})
\end{enumerate}

\section{Additional Experimental Details}
\label{appen:experimental details}
\subsection{Datasets}
\label{appen:datasets}
In this work, we categorize the datasets into \textit{in-context} and \textit{out-of-context} datasets. The videos from \textit{in-context} datasets consist of actions with frequent static context, \eg swimming in the swimming pool, while the videos from \textit{out-of-context} datasets contain actions occurring with an unusual static context, \eg dancing in the mall~\cite{choi2019can}.
We conduct the experiments on five \textit{in-context} benchmarks: Kinectics-400~\cite{k400} (K400), Kinectis-600~\cite{k600} (K600), UCF101~\cite{UCF101} (UCF), HMDB51~\cite{HMDB51} (HMDB), and Something-Something V2~\cite{ssv2} (SSv2). Additionally, we evaluate our approach on two \textit{out-of-context} benchmarks: SCUBA~\cite{li2023mitigating} and HAT~\cite{chung2022enabling}.

\noindent {\bf K400 and K600} are both comprehensive video datasets for human action recognition. K400 contains 400 action categories of approximately 240k training and 20k validation videos collected from YouTube, which covers a wide range of human actions, including sports activities, daily life actions, and various interactions, serving as a widely-used action recognition dataset for pre-training. The duration of video clips in K400 varies, with most clips being around 10 seconds long. This diversity in video duration helps models learn temporal dynamics and context for action recognition. K600 extends K400 by incorporating 220 additional new categories, thus enabling the evaluation of zero-shot learning capabilities on these novel categories. 

\noindent {\bf UCF} is a human action recognition dataset collected from YouTube, and consists of 13,320 video clips, which are classified into 101 categories. These 101 categories encompass a wide range of realistic actions including body motion, human-human interactions, human-object interactions, playing musical instruments and sports. Officially, there are three splits allocating 9,537 videos for training and 3,783 videos for testing.

\noindent {\bf HMDB} is a relatively small video dataset comprising a diverse range of sources, including movies, public databases, and YouTube videos, and is composed of 6,766 videos across 51 action categories (such as ``jump'', ``kiss'' and ``laugh''), ensuring at least 101 clips within each category. The original evaluation scheme employs three distinct training/testing splits, allocating 70 clips for training and 30 clips for testing of each category in each split.

\noindent {\bf SSv2} is a temporally focused video dataset across 174 fine-grained action categories, consisting of 168,913 training videos and 24,777 testing videos showing the objects and the actions performed on them. These action categories are presented using object-agnostic templates, such as ``Dropping [something] into [something]'' containing slots (``[something]'') that serve as placeholders for objects. This dataset focuses on basic, physical concepts rather than higher-level human activities, which challenges the temporal modeling capabilities.

\noindent {\bf SCUBA} is an out-of-distribution (OOD) video benchmark designed to quantitatively evaluate static bias in the background. It comprises synthetic out-of-context videos derived from the first test split of HMDB and UCF, as well as the validation set of K400. These videos are created by superimposing action regions from one video onto diverse scenes, including those from Place365~\cite{zhou2017places} and VQGAN-CLIP~\cite{crowson2022vqgan} generated scenes. 
Due to the differences in test sets and background sources, the domain gaps of SCUBA benchmarks vary. A domain gap is defined as the ratio of accuracies between the original test sets and synthetic datasets obtained by a 2D reference network, where a higher ratio indicates a greater domain gap with respect to static features.
The UCF-SCUBA and K400-SCUBA used in our experiments consist of 4,550 and 10,190 videos with domain gaps of $20.49$ and $6.09$, respectively, whose backgrounds are replaced by the test set of Place365.

\noindent {\bf HAT} is a more ``realistic-looking'' mixed-up benchmark for quantitative evaluation of the background bias by automatically generating synthetic counterfactual validation videos with different visual cues. It provides four Action-Swap sets with distinct characteristics: \textit{Random} and \textit{Same} refer to the swap of actions and backgrounds from different and same classes, respectively, while \textit{Close} and \textit{Far} denote the swap of videos from a class with similar and very different backgrounds, respectively. The UCF-HAT benchmark used in our experiments consists of Action-Swap videos in \textit{Close} and \textit{Far} sets from 5 closest and 30 farthest action categories, respectively, following the literature~\cite{chung2022enabling}. Note that we only consider videos from the first test split of UCF where all frames have human masks taking up 5\% to 50\% of the pixels to ensure that sufficient human and background cues are present in each generated Action-Swap video.

\subsection{Evaluation Protocols}
\label{appen:EP}
For the experimental settings, we follow the previous works~\cite{weng2023open,rasheed2023fine,ni2022expanding} for in-context generalization evaluations and perform the newly proposed out-of-context generalization evaluations described below.

\noindent {\bf In-context base-to-novel generalization.}
Under this setting, we divide the entire set of action categories into two equal halves: base and novel, with the most frequently occurring classes designated as the base classes. We conduct generalization evaluations on four in-context datasets, \ie K400, HMDB, UCF and SSv2, where the models are initially trained on the base classes within the training splits of the dataset, and evaluated on both base and novel classes within the validation splits. Every training split consists of 16 video clips of each base class. During inference within HMDB and UCF datasets, only the novel class samples in the first validation splits are used for evaluation. For K400 and SSv2 datasets, the full validation split of each is used for evaluation here. We report the results of the average top-1 accuracies for both base and novel classes as well as the harmonic mean.

\noindent {\bf In-context cross-dataset generalization.}
Under this setting, the models are fine-tuned on the training set of K400, and evaluated on three in-context cross-datasets, \ie UCF, HMDB and K600. We report top-1 average accuracies with performance variances on the three validation splits in case of UCF and HMDB. For K600, the models are evaluated on non-overlapping 220 categories with K400, and we report top-1 average accuracies over three randomly sampled splits of 160 categories.

\noindent {\bf Out-of-context cross-dataset generalization.}
Under the more challenging out-of-context cross-dataset setting, the models are also trained on K400, and then evaluated on two out-of-context datasets based on UCF, \ie UCF-SCUBA and UCF-HAT. We report the top-1 and top-5 average accuracies over the synthetic counterfactual validation splits from UCF's first validation split. We further conduct the closed-set out-of-context evaluation based on the K400-SCUBA benchmark and report the harmonic mean of the accuracies under in-context and out-of-context settings to comprehensively analyze the generalization of the models.

\subsection{Implementation Details}
\label{appen:implementation}
Each training video clip is sampled with $8$ frames uniformly, and each sampled frame is spatially scaled in the shorter side to $256$ pixels and is processed with basic augmentations like color jittering, random flipping and random cropping of $224\times 224$.
We leverage multi-view inference with $3$ temporal and $1$ spatial views per video and linearly aggregate the recognition results. 
For our Gaussian Weight Average scheme, we use $\mu=7$ and $\sigma^2=10$ for in-context base-to-novel generalization and $\mu=15$ and $\sigma^2=10$ for in-context and out-of-context cross-dataset generalization.
We also adopt decision aggregation with pre-trained CLIP with the video learner for in-context evaluations.
The experiments are conducted on two computing clusters with four NVIDIA RTX 24G 4090 GPUs.

\section{Additional Experimental Results}
\label{appen:ad-results}
\subsection{Additional Evaluations and Ablation Studies}
\label{appen:ad-ablation}

\begin{table}[!t]
\centering
\caption{Performance comparison (Top-1 Acc. (\%)) on HMDB dataset. We evaluate both in-context and out-of-context recognition (marked with $\star$) performances. We also report the harmonic mean (HM) of the results. $\ast$ and $\dagger$ indicate our implementation with frozen text learners.}
\label{tab:ooc(re)}
\resizebox{0.8\columnwidth}{!}{
\begin{tabular}{lccc}
\toprule
Method     & HMDB         & HMDB-SCUBA $\star$        & HM          
\\ \cmidrule{1-4}
X-CLIP     & 44.6 $\pm$ 5.2    & 22.5      & 31.0    \\
Open-VCLIP $\ast$ & \underline{53.8 $\pm$ 1.5} & \underline{25.9}  & \underline{35.0} \\
FROSTER $\dagger$    & 53.4 $\pm$ 1.2 & 23.7  &  32.8 \\
\rowcolor{gray!20}
\textbf{Ours} & \textbf{54.6 $\pm$ 1.1} & \textbf{32.5}  & \textbf{40.7} \\ \bottomrule
\end{tabular}
} 
\end{table}

\noindent {\bf Out-of-context cross-dataset evaluation on HMDB dataset.}
Regarding results shown in~\cref{tab:ooc(re)}, our method achieves the highest accuracy of $32.5\%$ on HMDB-SCUBA, and builds up an impregnable lead of $+5.7\%$ of HM results over the nearest competitor, enabling a superior balance for open-vocabulary generalization.

\begin{table}[]
\caption{Effect of the meta-optimization and Gaussian weight average (GWA) components in Open-MeDe. $\Delta$ denotes the performance gains of different schemes over the baseline. Our Open-MeDe is highlighted in \colorbox{gray!10}{gray}.}
\label{tab:ab}
\centering
\resizebox{0.85\columnwidth}{!}{
\begin{tabular}{cccccc}
\toprule
\begin{tabular}[c]{@{}c@{}}Meta\\ optimization\end{tabular} & GWA & UCF & $\Delta_{\text{UCF}}$ & UCF-SCUBA & $\Delta_{\text{UCF-SCUBA}}$ \\
\midrule
\NO & \NO & 78.5 & - & 28.3 & - \\
\YES & \NO & 83.2 & \Rc{$+$ 4.7} & 32.1 & \Ra{$+$ 3.8} \\
\NO & \YES & 82.3 & \Rc{$+$ 3.8} & 30.7 & \Ra{$+$ 2.4} \\
\rowcolor{gray!10}
\YES & \YES & \textbf{83.9} & \Rc{$+$ 5.4} & \textbf{33.5} & \Ra{$+$ 5.2} \\
\bottomrule
\end{tabular}
}
\end{table}
\noindent {\bf Effect of individual strategies in Open-MeDe.}
In~\cref{tab:ab}, we conduct ablation experiments to study effects of the core strategies in Open-MeDe \ie the cross-batch meta-optimization and GWA stabilization. Using only meta-optimization or GWA yields improvements of $+4.7\%/3.8\%$ and $+3.8\%/2.4\%$ over the plain learner on UCF / UCF-SCUBA, respectively. This indicates that meta-optimization substantially enhances generalization across both open-vocabulary settings in improving model's robustness, compared to GWA. 
These two components complement each other effectively, achieving substantial gains of $+5.4\%/5.2\%$.
Their integration leads to consistent improvements across diverse scenarios. 

\begin{table}[]
\caption{Effect of the learning rate $\delta$ for meta-optimization. We choose $\delta=1.67\times 10^{-3}$ as the default setting.}
\label{tab:delta}
\centering
\resizebox{0.8\columnwidth}{!}{
\begin{tabular}{ccccc}
\toprule
$\delta$ & UCF & HMDB & K600 & UCF-SCUBA \\ \cmidrule{1-5}
$1.67\times 10^{-1}$ & 83.7 & 54.3 & 73.5 & 33.2 \\
$1.67\times 10^{-2}$ & 83.7 & 54.5 & 73.6 & 33.4 \\
\rowcolor{gray!20}
$1.67\times 10^{-3}$ & 83.7 & 54.6 & 73.7 & 33.5 \\
$1.67\times 10^{-4}$ & 83.6 & 54.3 & 73.6 & 33.0 \\ \bottomrule
\end{tabular}
}
\end{table}
\noindent {\bf Effect of the learning rate $\delta$.}
As shown in~\cref{tab:delta}, we conduct experiments by setting the learning rate $\delta$ to different magnitudes.
It can be observed that as $\delta$ decreases, the general performance remains stable, which validates the robustness of our cross-batch meta-optimization. However, a further reduction to $1.67 \times 10^{-4}$ slightly decreases performance across most datasets, suggesting that the optimal value for $\delta$ lies at $1.67 \times 10^{-3}$, which is chosen as the default setting. This value achieves a balanced performance with the highest or nearly highest scores in each dataset, particularly noticeable on UCF-SCUBA benchmark.

\begin{table}[]
\caption{Effect of cross-batch meta-optimization.}
\label{tab:meta}
\centering
\resizebox{0.85\columnwidth}{!}{
\begin{tabular}{lcccc}
\toprule
Method & UCF & HMDB & K600 & UCF-SCUBA \\ \cmidrule{1-5}
Plain & 78.5 & 50.3 & 65.9 & 28.3 \\
Grad Accumulation & 78.9 & 50.5 & 66.5 & 28.9 \\
\rowcolor{gray!20}
\textbf{Meta Cross-batch} & \textbf{83.7} & \textbf{54.6} & \textbf{73.7} & \textbf{33.5} \\ \bottomrule
\end{tabular}
}
\end{table}
\noindent {\bf Effect of cross-batch meta-optimization.}
To investigate the efficacy of our cross-batch meta-optimization complementing the main paper, we further evaluate the performance using the scheme of gradient accumulation. To ensure a fair comparison of the total gradient steps with cross-batch meta-optimization, we accumulate the gradients over two steps before performing a single parameter update. As shown in~\cref{tab:meta}, the gradient accumulation technique demonstrates modest improvements over the plain method for both in-context and out-of-context benchmarks. This indicates that the strength of our meta-optimization approach lies in its ability to enhance known-to-open generalization, rather than doubling the batch size for a single parameter update.

\begin{table}[]
\caption{Effect of the randomness of the batch sampler for cross-batch meta-optimization. The ``\textit{similar}'' sampler denotes the usage of the most semantically similar classes across adjacent batches. We evaluate both in-context and out-of-context recognition (marked with $\star$) performances. HM: harmonic mean. Our default settings and results are highlighted in \colorbox{gray!20}{gray}.}
\label{tab:shuffle}
\centering
\resizebox{\columnwidth}{!}{
\begin{tabular}{lllll}
\toprule
Method &  Sampler & UCF (\%) & UCF-SCUBA$\star$ (\%) & HM (\%) \\ \cmidrule{1-5}
\multirow{2}{*}{Plain} & \textit{shuffle} & 78.5 & 28.3  & 41.6 \\
 &  \textit{initial}  & 77.7 (\Rb{$\downarrow$ 0.8})& 28.2 (\Rb{$\downarrow$ 0.1})& 41.4 (\Rb{$\downarrow$ 0.2})\\
\cmidrule{1-5}
\multirow{3}{*}{Meta Cross-batch} & \cellcolor{gray!10}\textbf{\textit{shuffle}} & \cellcolor{gray!10}\textbf{83.7} & \cellcolor{gray!10}\textbf{33.5}  & \cellcolor{gray!10}\textbf{47.8} \\
 & \textit{initial} & {82.5} (\Rb{$\downarrow$ 1.2}) & {28.9} (\Rb{$\downarrow$ 4.6}) & {42.8} (\Rb{$\downarrow$ 5.0}) \\
& \textit{similar} & {82.7} (\Rb{$\downarrow$ 1.0}) & {30.9} (\Rb{$\downarrow$ 2.6}) & {45.0} (\Rb{$\downarrow$ 2.8}) \\ \bottomrule
\end{tabular}
}
\end{table}
\noindent {\bf Effect of randomness of the batch sampler for cross-batch meta-optimization.}
To verify the efficacy of constructing tasks across batches with different inherent label distributions, we further conduct several additional studies about the sampling randomness during cross-batch meta-optimization. As shown in~\cref{tab:shuffle}, the randomness of the batch sampler is indeed an important factor to bring out the best of our cross-batch meta leaner, which improves the overall generalization greatly ($+5.0\%$ of harmonic mean) especially for out-of-context performance ($+4.6\%$ on UCF-SCUBA).
However, plain learner shows insensibility to the sampling randomness, experiencing negligible growth of generalization performance.
Without shuffling the batch sampler, our method still outperforms the non-shuffle plain learner by $+1.4\%$ of HM results.
By using the most semantically similar classes across support and query batches, it brings a relative performance \textit{decline} of $0.99\%$ and $2.64\%$ on UCF and UCF-SCUBA, respectively. We speculate that it amplifies inter-task semantic distribution shifts hindering cross-task generalization. In contrast, the default ensures consistent and balanced distributions of both inter- and intra-task variance.

\begin{table}[]
\caption{Effect of the batch size of tasks and samples for cross-batch meta-optimization. Our default settings and results are highlighted in \colorbox{gray!20}{gray}.}
\label{tab:batch}
\centering
\resizebox{0.85\columnwidth}{!}{
\begin{tabular}{cccccc}
\toprule
\multicolumn{2}{c}{Batchsize} & \multirow{2}{*}{UCF} & \multirow{2}{*}{HMDB} & \multirow{2}{*}{K600} & \multirow{2}{*}{UCF-SCUBA} \\ \cmidrule{1-2}
Task & Sample &  &  &  &  \\ \cmidrule{1-6}
2 & 8 & 83.5 & 54.3 & 73.2 & 33.5 \\
4 & 4 & 83.5 & 54.3 & 73.3 & 33.4 \\
\rowcolor{gray!20}
4 & 8 & 83.7 & 54.6 & 73.7 & 33.5 \\
4 & 16 & 83.8 & 54.6 & 73.9 & 33.6 \\
8 & 8 & 83.8 & 54.8 & 73.9 & 33.6 \\ \bottomrule
\end{tabular}
}
\end{table}

\noindent {\bf Effect of the batch size of tasks and samples.}
In~\cref{tab:batch}, we evaluate the performance with different batch sizes of the task and data for cross-batch meta-optimization. Each task consists of two data batches, one for the support set and one for the query set. From the results, we observe that increasing the batch size leads to slight improvements in performance, especially for K600. While larger batch sizes provide marginal improvements, they may not justify the increased computational cost. Thus, the default setting provides an effective balance between performance and computational efficiency.

\begin{table}[]
\caption{Effect of the CLIP ensemble. We evaluate the performance of integrating the CLIP ensemble within the weight and decision spaces. \textit{Naive} denotes applying only the video learners for evaluations without further CLIP ensemble.}
\label{tab:ensemble}
\centering
\resizebox{\columnwidth}{!}{
\begin{tabular}{llcccc}
\toprule
Method & CLIP ensemble & UCF & HMDB & K600 & UCF-SCUBA \\ \cmidrule{1-6}
\multirow{3}{*}{VCLIP} & Naive & 78.5 & 50.3 & 65.9 & 28.3 \\
 & Weight & 80.1 & 51.9 & 71.0 & 26.6 \\
 & Prediction & 80.3 & 52.1 & 71.2 & 27.0 \\ \cmidrule{1-6}
\multirow{3}{*}{Open-VCLIP} & Naive & 81.4 & 53.2 & 71.5 & 30.0 \\
 & Weight & 83.3 & 53.8 & 73.0 & 28.9 \\
 & Prediction & 83.4 & 54.0 & 73.2 & 29.9 \\ \cmidrule{1-6}
\multirow{3}{*}{\textbf{Open-MeDe}} & Naive & 83.3 & 54.3 & 73.5 & \cellcolor{gray!20}\textbf{33.5} \\
 & Weight & 83.6 & 54.4 & 73.6 & 29.9 \\
 & \cellcolor{gray!20}Prediction & \cellcolor{gray!20}\textbf{83.7} & \cellcolor{gray!20}\textbf{54.6} & \cellcolor{gray!20}\textbf{73.7} & 32.0 \\ \bottomrule
\end{tabular}
}
\end{table}
\noindent {\bf Effect of the CLIP ensemble.}
In~\cref{tab:ensemble}, we evaluate the effectiveness of the CLIP ensemble in the weight space and decision space, with the ensemble ratios all set to $0.5$.
The results demonstrate that both types of CLIP ensemble improve performance in in-context evaluations, with the prediction-based ensemble yielding the most consistent gains across all methods. This suggests that integrating CLIP predictions effectively leverages the strengths of CLIP, leading to significant performance enhancements, particularly over the naive approach. However, there is a noticeable drop on UCF-SCUBA for the out-of-context generalization, indicating that the static generalization derived from the CLIP ensemble can adversely affect the model's robustness and overall generalizability.

\begin{table}[]
\caption{Effect of mitigating static bias in action recognition with various training strategies. We report the Top-1 Acc. (\%) and harmonic mean (HM) of both in-context  (IC) and out-of-context (OC) generalization performance for closed-set and zero-shot action recognition. \ding{55} indicates that the methods are not capable of zero-shot action recognition.}
\label{tab:debias}
\centering
\resizebox{\columnwidth}{!}{
\begin{tabular}{lllcccccc}
\toprule
\multirow{2}{*}{Method} & \multirow{2}{*}{Pretrain} & \multirow{2}{*}{Training Strategy} & \multicolumn{3}{c}{K400 (closed-set)} & \multicolumn{3}{c}{UCF (zero-shot)} \\
 &  &  & IC & OC & HM & IC & OC & HM \\ \cmidrule{1-9}
BE~\cite{wang2021removing} & ImageNet & Debiasing & 73.9 & 41.9 & 53.5 & \ding{55} &\ding{55}&\ding{55}\\
FAME~\cite{ding2022motion} & K400 & Debiasing & 73.8 & \underline{49.0} & 58.9 &\ding{55}&\ding{55}&\ding{55}\\
StillMix~\cite{li2023mitigating} & ImageNet & Debiasing & 73.9 & 43.4 & 54.7 &\ding{55}&\ding{55}&\ding{55}\\
DEVIAS~\cite{bae2023devias} & VideoMAE & Disentangle & 77.3 & \textbf{51.8} & \textbf{62.0} &\ding{55}&\ding{55}&\ding{55}\\
VCLIP & CLIP & Plain & \underline{80.1} & 42.4 & 55.4 & 78.5 & 28.3 & 41.6 \\
\rowcolor{gray!20}
\textbf{Open-MeDe} & CLIP & Meta-optimization & \textbf{81.5} & 46.6 & \underline{59.3} & \textbf{83.9} & \textbf{33.5} & \textbf{47.9}
\\ \bottomrule
\end{tabular}
}
\end{table}
\noindent {\bf Effect of static debiasing strategies.}
In~\cref{tab:debias}, we compare Open-MeDe with several baselines especially designed for mitigating static bias in action recognition, including three scene-debiasing methods (BE~\cite{wang2021removing}, FAME~\cite{ding2022motion} and StillMix~\cite{li2023mitigating}) and a state-of-the-art action-scene disentanglement method (DEVIAS~\cite{bae2023devias}). Note that DEVIAS leverages additional scene labels for disentangled video representation. As can be seen from the results, while FAME and DEVIAS perform well in the K400 closed-set out-of-context evaluation against static bias, they fall short in in-context performance and lack zero-shot inference capability. In contrast, our Open-MeDe, despite not employing explicit debiasing or disentangled action modeling, achieves favorable out-of-context generalization with a balanced harmonic mean. This highlights its robust generalizability across both in-context and out-of-context scenarios, particularly excelling in zero-shot generalization.

\begin{figure}[!t]
    \centering
    \small
    \includegraphics[width=\columnwidth]{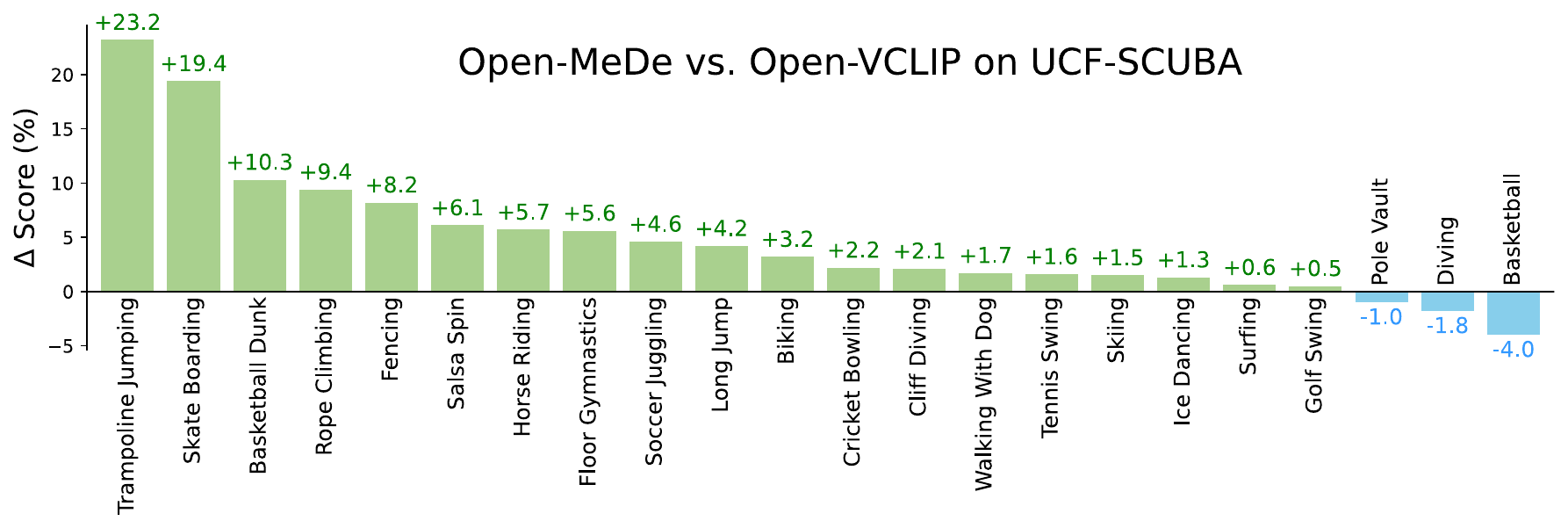}
    \caption{Comparison between Open-MeDe and OpenVCLIP across $22$ classes on UCF-SCUBA.}
    \label{fig:classes}
\end{figure}

\noindent {\bf Analysis of class-wise performance.}
In~\cref{fig:classes}, we further present the improvements of our Open-MeDe over OpenVCLIP on out-of-context UCF-SCUBA across 22 novel classes.
It can be observed that Open-MeDe wins across $19$ of the $22$ classes. The improved categories involve localized motions, where most of the static content is misinterpreted by irrelevant context noise on UCF-SCUBA. We attribute these gains primarily to its effectiveness in static debiasing and capturing fine-grained dynamics. However, its performance is slightly compromised in cases involving team sports or rapid shifts in spatial locations.  

\begin{table}[]
\caption{Comparison of the training cost. We report the results of K400 training on four GPUs (24G RTX 4090). We maintain an equal batch size of 8 videos per GPU across all models. }
\label{tab:train}
\centering
\resizebox{\columnwidth}{!}{
\begin{tabular}{lcccc}
\toprule
Method & Params (M) & FLOPs (G) & CUDA mem. (GB) & Epoch time (min) \\ \cmidrule{1-5}
VCLIP & 149.62 & 152.11 & 14.14 & 110.10 \\
Open-VCLIP & 149.62 & 152.11 & 20.09 & 109.26 \\
FROSTER & 299.77 & 152.11 & 21.07 & 80.95 \\
\rowcolor{gray!20}
\textbf{Open-MeDe} & 149.62 & 152.11 & 16.59 & 74.33 \\ \bottomrule
\end{tabular}
}
\end{table}
\subsection{Training cost analysis}
In~\cref{tab:train}, we show the training cost analysis of our approach and compare it with other methods under identical training conditions. All approaches utilize the same video learner, ensuring equal GFLOPs. Our Open-MeDe achieves the lowest CUDA memory usage at $16.59$ GB and a significantly reduced epoch time of $74.33$ minutes, compared to other methods. This demonstrates its efficiency in terms of training time and memory consumption, providing a cost-effective solution without compromising on computational complexity.

\subsection{Visualization Results}
\begin{figure*}[t]
    \centering
    \includegraphics[width=\textwidth]{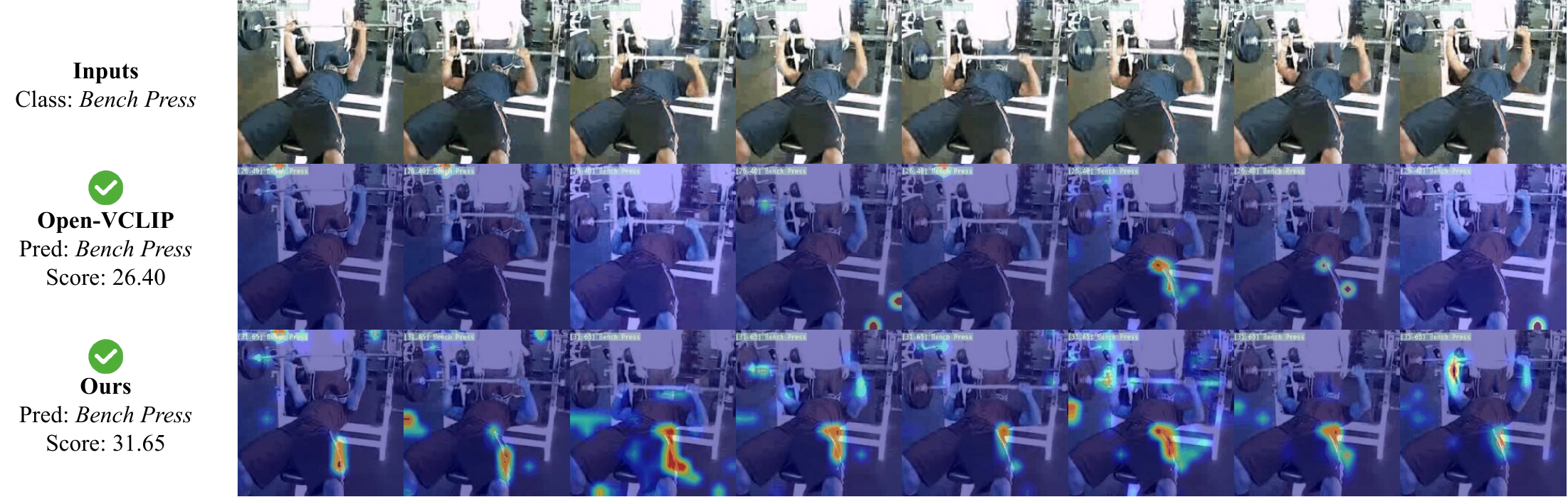}
    \caption{Visualizations of attention maps and predictions for ``\textit{Bench Press}'' in the in-context setting. Both Open-VCLIP and our proposed framework correctly predict the action, while ours achieves a higher score. Additionally, our framework demonstrates enhanced attention to the key elements associated with the action, which highlights its effectiveness in capturing nuanced and discriminative features, leading to more confident predictions.}
    \label{fig:benchpress}
\end{figure*}

\begin{figure*}[t]
    \centering
    \includegraphics[width=\textwidth]{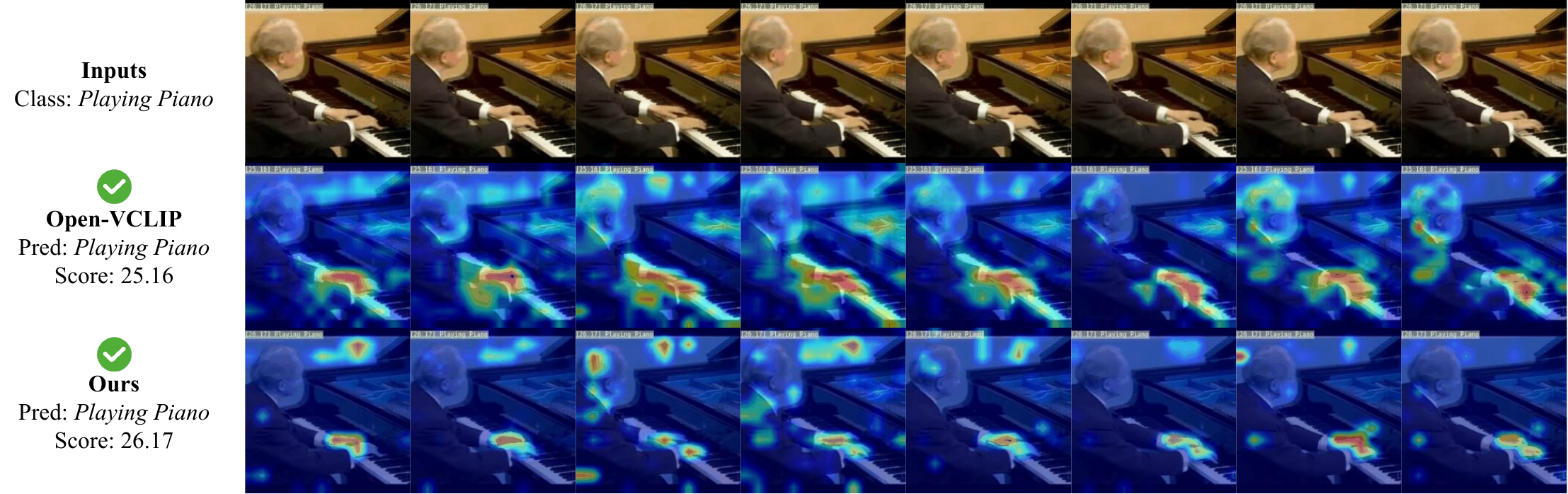}
    \caption{Visualizations of attention maps and predictions for ``\textit{Playing Piano}'' in the in-context setting. Our method emphasizes the subtle movements of the action rather than redundant visual appearances, demonstrating its effectiveness of capturing critical motion cues.}
    \label{fig:playingpiano}
\end{figure*}

\begin{figure*}[t]
    \centering
    \includegraphics[width=\textwidth]{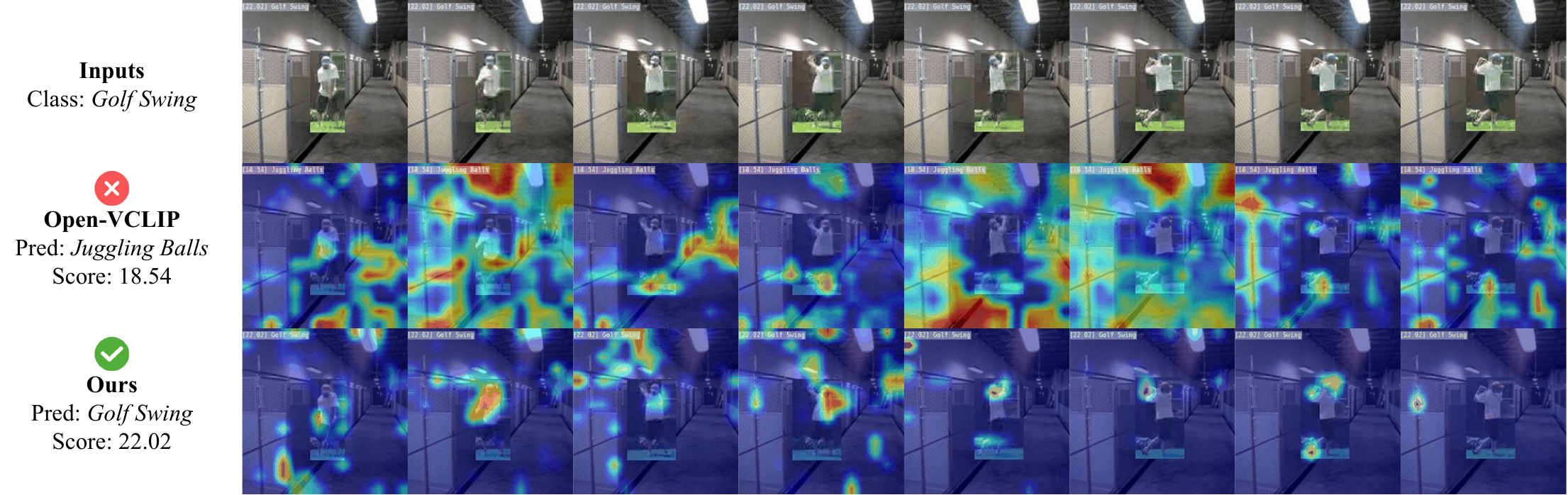}
        \caption{Visualizations of attention maps and predictions for ``\textit{Golf Swing}'' in the out-of-context setting. Our method successfully classifies the action and effectively captures key visual cues associated with the motion, demonstrating the improved robustness. However, Open-VCLIP misclassifies the action as ``\textit{Juggling Balls}'' due to its large static bias.}

    \label{fig:golf}
\end{figure*}

\begin{figure*}[t]
    \centering
    \includegraphics[width=\textwidth]{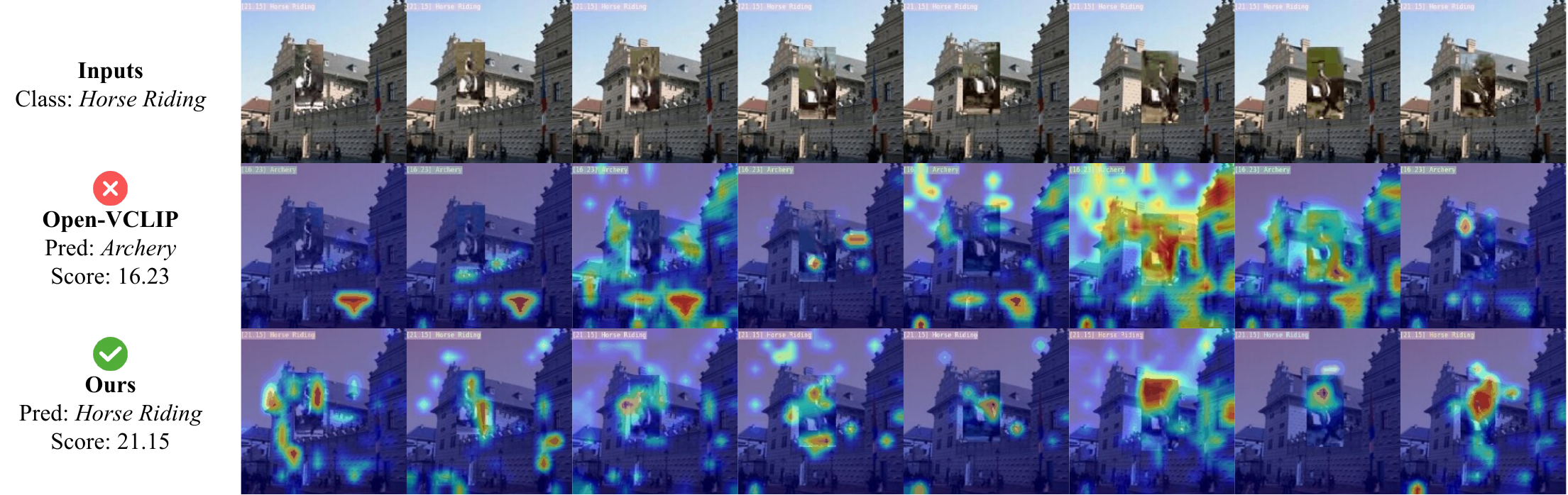}
    \caption{Visualizations of attention maps and predictions for ``\textit{Horse Riding}'' in the out-of-context setting. Our method outperforms Open-VCLIP by accurately attending to critical dynamic information specific to the true action, showcasing its robustness and reliability in discerning action-relevant features under challenging out-of-context scenarios.}
    \label{fig:horseriding}
\end{figure*}

\begin{figure*}[t]
    \centering
    \includegraphics[width=\textwidth]{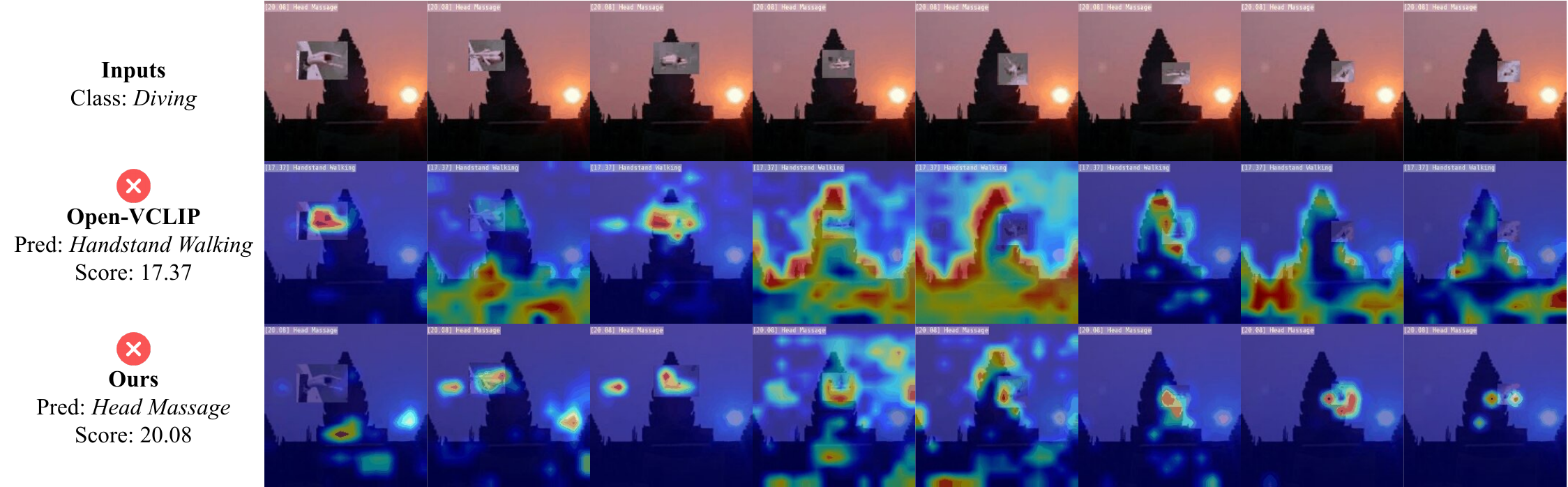}
    \caption{Visualizations of attention maps and predictions for ``\textit{Diving}'' in the out-of-context setting. Both methods struggle to classify the action correctly, suggesting more room for improvement under this challenging scenario. Despite the incorrect prediction, our method reflects a better focus on motion-relevant areas, which indicates its effectiveness of mitigating static bias. }
    \label{fig:diving}
\end{figure*}

As shown in~\cref{fig:benchpress,fig:playingpiano,fig:golf,fig:horseriding,fig:diving}, we present additional visualization comparisons of Open-VCLIP and the proposed framework under in-context and out-of-context scenarios. Overall, our approach effectively attends to more motion-relevant regions, achieving higher confidence scores and correct predictions in most cases. This demonstrates its greater reliability, and robust generalizability in open-vocabulary action recognition tasks.
\section{Discussions}
\label{appen:discussion}
In this part, we further elucidate the core distinction between the proposed method and similar paradigms through comparative analysis.

\noindent {\bf Meta learner \vs Plain learner.}
As discussed in the main paper, Open-MeDe formulates the video learner into a meta learner by employing the cross-batch meta-optimization scheme that mimics sequences of known-to-open generalization tasks, enhancing adaptability to unseen data through iterative virtual evaluations during training.
Plain learners, such as those employing standard fine-tuning paradigms on CLIP-based video learners, are typically straightforward and focus on in-distribution class-specific knowledge.
Following a traditional gradient descent over a single objective function can lead to a narrower optimization landscape prone to overfitting.
Therefore, plain learners can gain reasonable in-context performance but struggle to generalize to novel and out-of-context scenarios due to the tendency to overfit in training data and short-cutting static cues. 

In contrast, our meta learner is designed to derive the training towards learning more generalizable features by optimizing not just for class-specific knowledge but for adaptability across diverse known-to-open tasks. 
It explicitly counteracts inherent known and static biases by leveraging feedback from virtual evaluations, ensuring the video learner does not over-rely on vulnerable static cues.
By alternating between \textit{meta training} (\wrt support data) and \textit{meta testing} (\wrt query data), the meta learner ensures smoother optimization trajectories and enhanced robustness in a cost-effective manner.
The episodic training of the meta learner fosters adaptability across varying class distributions, making it highly effective for open-vocabulary tasks.


\noindent {\bf Meta-optimization \vs Train-validation.}
In our meta-optimization framework, training involves two key stages: \textit{meta training} (on support data) and \textit{meta testing} (on query data). The query data evaluation provides generalization feedback via loss gradients, enabling the learner to adjust the learning trajectory to prioritize generalizable features. This iterative approach inherently targets learning to generalize and mitigates overfitting by encouraging robust learning across diverse distribution shifts.
Conversely, the train-validation paradigm typically partitions data into training and validation subsets, optimizing model parameters by minimizing errors on the training data while evaluating performance on a held-out validation set for hyper-parameter tuning or early stopping.
This paradigm monitors the generalization performance indirectly by balancing the performance between training and validation data without explicitly improving the open-vocabulary generalization capability toward novel data.

Both paradigms leverage the feedback to refine model training, where the feedback of meta-optimization comes from query evaluations, while in train-validation, it arises from validation performance.  
Additionally, the feedback of train-validation is aggregated at coarser intervals, limited to hyper-parameter adjustment on constant training-validation splits.
It is worth noting that the meta-optimization provides granular, iterative feedback during training, manifesting as loss gradients to refine generalizable representation learning by dynamically constructing tasks with support-query splits.
Therefore, the proposed meta-optimization framework provides a more robust and explicit mechanism for adapting to novel data, setting a new baseline for open-vocabulary action recognition.


\noindent {\bf Cross-batch meta-optimization \vs Gradient accumulation.}
As introduced in the Open-MeDe framework, the proposed cross-batch meta-optimization takes inspiration from meta-learning with minimal modification to the standard training setup, which leverages adjacent mini-batches in training, treating one as the support batch (\textit{meta training}) and the subsequent as the query batch (\textit{meta testing}).
It aims to explicitly promote generalization by evaluating how well the model can adapt its learned parameters to open or dynamically different data distributions, thereby mitigating inherent and static biases in the video learner.
When it comes to the gradient accumulation technique, by simulating large batch training, it aggregates gradients over multiple mini-batches and applies the update after a predefined number of steps, emphasizing the efficiency of stabilizing training and improving convergence on hardware-constrained scenarios. However, it primarily improves training stability without inherently targeting adaptability and enhanced generalization.
Therefore, cross-batch meta-optimization differs fundamentally from gradient accumulation in its goal and methodology, which achieves a superior balance between specialization and generalization.

\noindent {\bf Meta-debiasing with MVSGG~\cite{xu2022meta}.}
1) Objective \wrt mitigating biases.
MVSGG addresses certain conditional biases within video scene generation tasks, targeting long-tailed data issues.
Here, we tackle a ubiquitous challenge for video understanding, \ie mitigating static bias present in video learners.
2) Methodology \wrt meta-optimization.
MVSGG emphasizes on constructing various types of conditional biases within data at each training epoch, with its meta-optimization employed per epoch against specific biases.
We perform meta-optimization densely in iterations with a diverse distribution of tasks. The evaluation on a subsequent batch explicitly simulates known-to-open generalization and mitigates static bias implicitly.
3) Application scope \wrt generalization.
MVSGG enhances model's generalization under closed-set settings against conditional biases within training data.
Notably, we achieve more robust open-vocabulary generalization beyond training data, where MVSGG is insufficient to our requirements. 
4) Computational cost \wrt task construction.
MVSGG requires careful organization of training data, significantly increasing computational cost. Remarkably, our method incurs no additional computational overhead compared to standard training by effortlessly utilizing cross-batch data.

\noindent {\bf Gaussian self-ensemble with PromptSRC~\cite{khattak2023self}.}
Our GWA is related to PromptSRC with two key differences:
1) Objective \wrt implementation.
We aim to achieve a generic optimal solution for video learners by assigning different weights to learner's parameters during optimization, while PromptSRC focuses on regularizing prompt learning to reduce overfitting with frozen backbones.
2) Patching strategy \wrt start point.
Our GWA starts after fine-tuning the pre-trained weights of the learner (\eg, CLIP weights), which exhibits substantial static-related knowledge. With the purpose of mitigating static bias, the initial patching weights are sampled from low Gaussian probabilities.
However, the start point of PromptSRC is randomly initialized, given the prompt learning framework, where lower weight assignments guarantee the task-specific knowledge. 

\section{Broader Impacts and Limitations}
\label{appen:limitation}

\noindent {\bf Broader Impacts.}
The proposed Open-MeDe framework for open-vocabulary action recognition introduces substantial advancements in several key aspects, underscoring its broader impact on both research and real-world applications: 
1) By addressing the overfitting to static cues inherent in pre-trained models like CLIP, Open-MeDe introduces innovative solutions for robust generalization. Its combination of meta-optimization and Gaussian self-ensemble stabilization enables robust performance in challenging out-of-context scenarios, providing a pathway for video learners to bridge the gap between image and video modalities effectively. 
2) Unlike previous approaches reliant on CLIP regularization, Open-MeDe reduces computational overhead and efficiently balances class-specific learning with generalization capabilities, leveraging a cross-batch meta-optimization approach. 
3) Open-MeDe demonstrates remarkable adaptability across diverse scenarios, including base-to-novel, cross-dataset, and out-of-context evaluations. Its model-agnostic design enables seamless integration with various CLIP-based video learners, enhancing performance across parameter-efficient fine-tuned, partially-tuned, and fully-tuned video learners. This flexibility significantly broadens its utility, making it a versatile tool for tasks requiring robust generalization without extensive domain-specific tailoring.
4) Extensive experiments demonstrate the state-of-the-art results achieved by our Open-MeDe, highlighting its role in advancing general video understanding. Our framework can empower many downstream applications, such as video-based surveillance and security, autonomous vehicles, human-computer interaction, \etc.

\noindent {\bf Limitations.}
Despite achieving promising open-vocabulary generalization with our framework, the out-of-context scenarios remain challenging and constrained by the reliance on temporal and static feature alignment. 
Specifically, scenarios with extreme domain shifts (\eg, SCUBA and HAT benchmarks) show significant performance gaps. 
However, the residual influence of static visual cues persists, particularly in complex video backgrounds and more compact foregrounds.
Incorporating stronger, explicitly targeted debiasing strategies, such as adversarial learning or counterfactual data augmentation, may further enhance robustness, which will be explored in our future work.

\clearpage

\end{document}